\documentclass[11pt]{article}

\usepackage[preprint]{acl}

\usepackage{times}
\usepackage{latexsym}
\usepackage{hyperref}
\usepackage{xcolor}

\usepackage[T1]{fontenc}
\usepackage[utf8]{inputenc}

\usepackage{microtype}

\usepackage{inconsolata}

\usepackage{graphicx}

\usepackage{url}
\usepackage{multirow}
\usepackage{enumitem}
\usepackage{graphicx}
\usepackage{subcaption}
\usepackage{multicol,lipsum}

\usepackage{booktabs, multirow} % for borders and merged ranges
\usepackage{soul}% for underlines
\usepackage{xcolor,colortbl} % for cell colors
\usepackage{changepage,threeparttable} % for wide tables
\usepackage{makecell}

\usepackage{amsmath}
\usepackage{xspace}
\usepackage{siunitx}

\usepackage[utf8]{inputenc}
\usepackage{booktabs}
\usepackage{tabularx}
\usepackage{url}
\usepackage{hyperref}
\usepackage{fontspec}
\newfontfamily{\multilingualfont}{FreeSerif.otf}
\newcommand{\expRaw}{\textbf{raw\_wiki}\xspace}
\newcommand{\expPrimary}{\textbf{+filters}\xspace}

\newcommand{\expRandom}{\textbf{random}\xspace}

\title{How Good is Your Wikipedia? \\Auditing Data Quality for Low-resource and Multilingual NLP}

\author{\normalsize Kushal Tatariya$^{1*}$ \ \ Artur Kulmizev$^{2*}$ \ \ Wessel Poelman$^{1}$ \ \
Esther Ploeger$^{3}$ \\\normalsize \textbf{Marcel Bollmann}$^{4}$ \ \ \textbf{Johannes Bjerva}$^{3}$ \ \
\textbf{Jiaming Luo}$^{5}$ \ \ \textbf{Heather Lent}$^{3}$ \ \ \textbf{Miryam de Lhoneux}$^{1}$\\
      \normalsize   $^1$Department of Computer Science, KU Leuven  \ \  $^2$CENTAL, UCLouvain  \\
       \normalsize   $^3$Department of Computer Science, Aalborg University \\
        \normalsize  $^4$Department of Computer and Information Science, Linköping University \\
         \normalsize $^5$Google Translate \\
           }

\begin{document}
\maketitle

\begin{abstract}
Wikipedia's perceived high quality and broad language coverage have established it as a fundamental resource in NLP. However, in recent years, such assumptions of high quality have become the subject of scrutiny in low-resource and multilingual contexts. In this study, we subject the entirety of non-English Wikipedia to a data filtering procedure typically reserved for noisy web-text --- a process which removes a large percentage of the collection's data. In analysing the removed data, we reveal numerous systematic quality issues, such as script and language contamination, repeated template and placeholder articles, and a high concentration of bot-generated content. We consolidate these findings into a 4-level quality ranking of Wikipedia, which shows strong correspondence with alternative quality measures and heuristics. Lastly, we evaluate the downstream impact of quality filtering in three practical language modelling scenarios, showing that models trained on filtered data largely match or outperform those trained on raw Wikipedia, with the largest gains observed for lower-quality language editions. Ultimately, our experiments serve as a first step in establishing quality-aware best practices for Wikipedia utilization in NLP, laying groundwork that can inform future dataset creation and curation efforts.
\end{abstract}

\section{Introduction}
\begingroup\def\thefootnote{*}\footnotetext{Equal contribution}\endgroup
Wikipedia has long served as an invaluable NLP resource, representing an international collaborative effort in documenting an ever-changing world. Its open-source framework --- through which authors, editors, and administrators cooperate to continuously refine and expand content --- ensures that information contained therein is not only current, but also well-written, accurate, and relevant to any given subject. For these reasons, Wikipedia has been employed for a myriad of NLP applications, such as language model (LM) training and benchmarking \citep{merity2016pointer}, question-answering \citep{hewlett-etal-2016-wikireading,rajpurkar-etal-2016-squad,joshi-etal-2017-triviaqa} and knowledge-base creation \citep{10.1145/2629489,lehmann-etal-2024-beyond}, among many others. Wikipedia is also a staple resource in multilingual NLP, where it is commonly employed for pretraining multilingual LMs \citep{devlin-etal-2019-bert,conneauxlm100} and creating benchmarks for downstream tasks such as named-entity recognition \cite{rahimi-etal-2019-massively} and machine translation \cite{schwenk-etal-2021-wikimatrix}. 

Recently, increasing attention to data quality has led to the adoption of various commonplace data filtering practices within NLP, whereby noisy web-scale data is distilled into a more salient signal for LM pretraining. Wikipedia has often been employed as a `high quality' reference in this process \citep[\textit{e.g.,}][]{wenzek-etal-2020-ccnet}, largely due to its encyclopaedic domain and culture of rigorous community moderation. However, the assumption of Wikipedia's inherent `high quality' --- though generally reliable for English and select high-resource languages --- is now increasingly being challenged in the multilingual setting \cite{kreutzer-etal-2022-quality}. Indeed, it has been found that Wikipedias for certain lower-resourced languages can contain unnatural, inorganic, and machine-translated text, among other noise, which can be incomprehensible to native speakers \cite{alabi-etal-2020-massive, lent2024creoleval}. As a result, poor data quality for such languages typically leads to poor performance and untrustworthy models, ultimately providing unusable technologies for speakers \cite{held2023materiallenscolonialitynlp, nicholas2023losttranslationlargelanguage, durmus2024measuringrepresentationsubjectiveglobal}. 

We critically examine the extent of Wikipedia's `high quality' in a non-English setting in this study. Estimating data quality via linguistic analyses and native speaker judgments is a costly endeavour for individual Wikipedias, and not feasible across the collection's 340+ active languages with an estimated 65 million articles. Thus, we operationalise existing automatic data cleaning methods to reliably determine Wikipedia quality in a data-driven manner. Specifically, we employ two distinct data cleaning processes --- \textit{script filtering} and \textit{deduplication} --- and apply them successively to every available Wikipedia, excluding English. Given that these techniques are typically applied to noisy, uncurated web-text, the extent to which a given Wikipedia is robust to such operations serves as an inverse measure of its `quality'. We demonstrate that each cleaning step is able to target distinct categories of noise, revealing systematic differences in content quality across language editions --- regardless of size. Furthermore, we find that the downstream performance of models trained on `cleaned' Wikipedias is often preserved --- and sometimes improved --- with respect to the original, unfiltered data. Overall, our contributions are as follows:

\begin{enumerate}[noitemsep,topsep=1pt]
    \item We present empirical evidence of quality issues across Wikipedia, establishing a quality taxonomy that can be employed for future research.
    \item We perform downstream evaluation of filtered Wikipedias across three practical low-resource and multilingual scenarios. 
    \item We release an open-source NLP toolkit for cleaning text data in many languages.\footnote{\url{https://github.com/akulmizev/texieve}}
\end{enumerate}

\begin{figure*}[htb!]
    \centering
    \includegraphics[width=\linewidth]{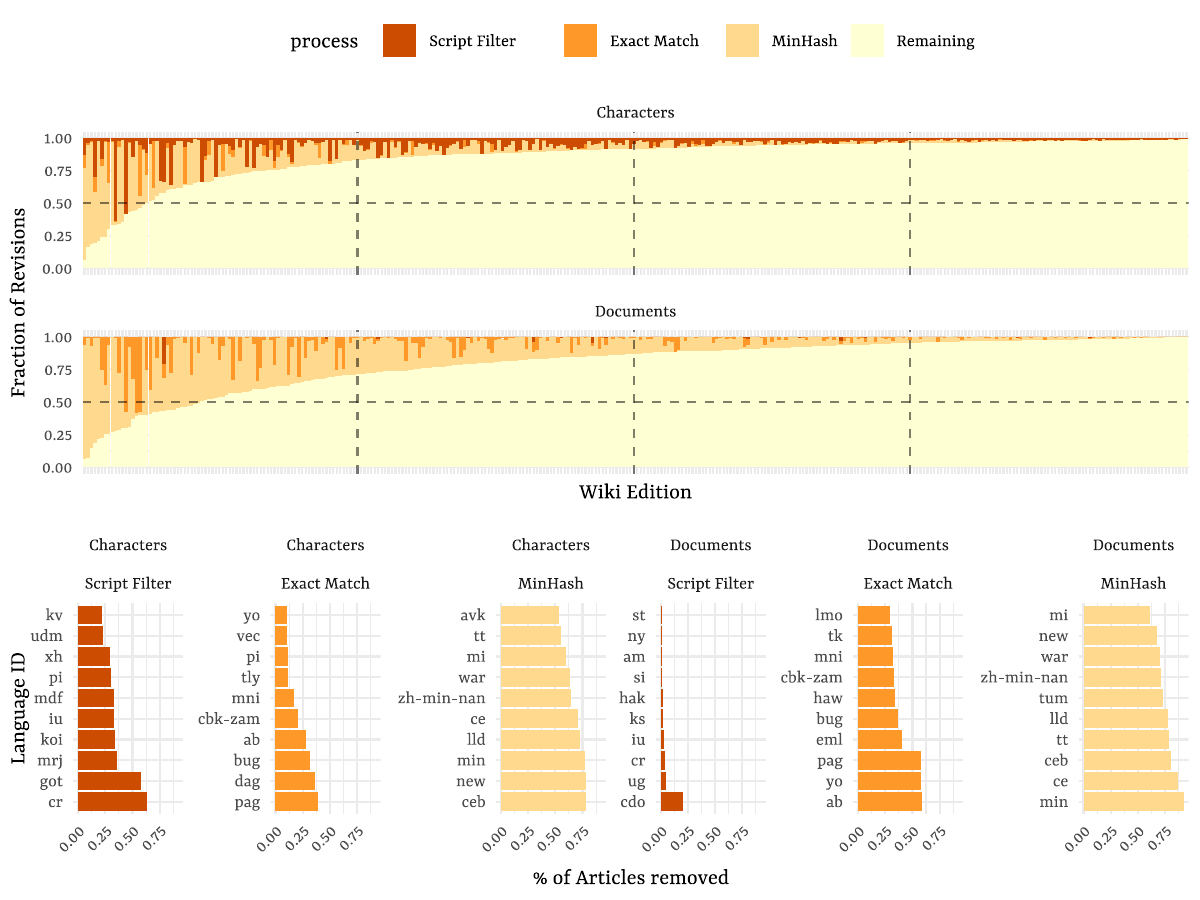}
    \caption{\footnotesize \textit{Top:} Fraction of data removed by each filter, sorted by remaining data in each Wikipedia. \textit{Bottom:} Top 10 Wikipedias according to fraction of data removed by each filter.}
    \label{fig:main}
\end{figure*}

\section{Related Work}

\subsection{Data Filtering}\label{rw:data-filtering}
Recent studies in large language modelling have demonstrated the benefits of filtering web-scale data \cite{wenzek-etal-2020-ccnet, raffel2023exploringlimitstransferlearning, penedo2024fineweb, ai2024yiopenfoundationmodels}. 
In this context, data filtering typically involves a combination of techniques, ranging from basic document-level heuristics to corpus-level statistics that serve as quality proxies \citep{albalak2024surveydataselectionlanguage}. For example, the popular C4 dataset includes heuristics that remove documents containing code (\textit{e.g.}, `Javascript'), boiler plate legal policies (\textit{e.g.}, `Terms and Conditions'), or inappropriate terms, among other hand-crafted rules \cite{raffel2023exploringlimitstransferlearning}. Alongside heuristics, \textit{deduplication} is another common filtering technique designed to remove duplicate documents from a given dataset, only keeping one copy thereof. A particularly popular algorithm is MinHash, due to its effectiveness in removing semantically identical documents, as well as its scalability across CPU nodes \cite{penedo2024fineweb}. MinHash has likewise been combined with downstream clustering-based deduplication to filter documents in a more aggressive manner \cite{tirumala2023d4improvingllmpretraining}. More sophisticated techniques have recently also been employed: for example, \citet{ai2024yiopenfoundationmodels} leverage learned filters (\textit{i.e.}, quality classifiers and cluster-based filters), which are designed to preserve documents resembling a `high quality' reference corpus -- typically Wikipedia \cite{wenzek-etal-2020-ccnet, together2023redpajama}.

%%%%%%%% Do we want to talk about Wikipedia-specific filtering techniques ??????? 
% Finally, for Wikipedia filtering specifically, previous approaches towards the automatic assessment of Wikipedia quality can be divided into two categories: metadata-based approaches and content-based approaches. Within the former, ... make use of information from author authority, citations, edit numbers, etc. 
% Work in this space has aimed to evaluate the quality of individual pages on Wikipedia for tasks such as vandalism detection \cite{MARTINEZRICO2019248}.
% A note on languages -- many works are on English, some german, but little work has been done beyond these languages. Say something about Lewoniewski et al. (2017, 2018, 2019, 2020) ...

\subsection{Wikipedia in Multilingual NLP}\label{rw:wiki-in-lr-nlp}
Wikipedia's widespread language coverage and reputation as a high-quality data trove has made it a vital resource in multilingual NLP applications. It has served as the backbone of many multilingual benchmark datasets, such as
\textsc{WikiAnn} \cite{pan-etal-2017-cross},
\textsc{XQuAD}  \cite{artetxe-etal-2020-cross},
\textsc{MLQA} \cite{lewis-etal-2020-mlqa}, 
\textsc{TyDiQA-GoldP} \cite{clark-etal-2020-tydi}, 
and BUCC \cite{zweigenbaum-etal-2017-overview}, %All these datasets are part of XTREME. 
as well as a common source of pretraining data for popular multilingual LMs, such as mBERT \cite{devlin-etal-2019-bert} and XLM \cite{Lample2019CrosslingualLM}. Beyond this, it was notably employed as the training data for \textsc{CCNet}'s language identification classifier, which is primarily used for partitioning the popular Common Crawl dataset along language boundaries. Given that Common Crawl serves as the basis for many contemporary open-source pretraining datasets, the quality and performance of such a classifier becomes paramount in retaining relevant data. 

As a consequence of its ubiquity in NLP, Wikipedia has recently attracted increased scrutiny for its data quality. For example, \citet{kreutzer-etal-2022-quality} find that the Wikipedia-oriented \textsc{WikiMatrix} \cite{schwenk-etal-2021-wikimatrix} contains a number of deprecated, incorrect, or otherwise mislabelled language codes, among the 86 languages included \cite{kreutzer-etal-2022-quality}. 
% Their additional manual audit of crawled documents uncovers that quality gradually declines in correlation with resourcedness. 
\citet{lignos-etal-2022-toward} identify similar shortcomings for lower-resourced languages in \textsc{WikiAnn} and even Wikidata.\footnote{\url{https://wikidata.org}} Studies on selected low-resource languages likewise confirm such findings. 
For example, \citet{alabi-etal-2020-massive} observe that the Yorùbá and Twi Wikipedias are ostensibly written by non-native speakers and lack the necessary diacritics and variation to be relevant to speakers, while \citet{lent2024creoleval} find that sentences from Wikipedias for Creole languages are often incomprehensible by speakers and require manual correction. 

\subsection{Data Quality versus Data Quantity}\label{rw:dq-vs-dq}
The notion of data \textit{quality} is inherently subjective. 
Quantitatively, it is commonly measured by performance gains on a downstream NLP task. 
Qualitatively, it can be understood to signify utterances that are accepted as valid by proficient speakers of a language, or, alternatively, text samples that do not promote toxic or offensive sentiments \cite{kreutzer-etal-2022-quality, van-noord-etal-2024-language}.  
Such definitions can sometimes be at odds with one another. For example, \citet{longpre-etal-2024-pretrainers} show that a smaller, carefully filtered dataset can improve performance on downstream tasks in contrast to a larger, unfiltered dataset, but that the resulting model becomes more likely to generate toxic content. 
Likewise, diminishing returns have been demonstrated from training on ever-increasing amounts of data \cite{conll-2023-babylm}. 
For example, language models pretrained on C4 (745GB) marginally outperform those trained on Wikipedia (16GB) over benchmarks like GLUE \citep{wang-etal-2018-glue}, SuperGLUE \citep{wang2019superglue}, and SQuAD \citep{rajpurkar-etal-2016-squad}. 
However, many of these benchmarks were sourced from Wikipedia itself, suggesting that quantity is less important than domain \cite{raffel2023exploringlimitstransferlearning}. 

\begin{table*}[htbp!]
    \centering
    \scriptsize
    \multilingualfont
    \begin{tabularx}{\textwidth}{l X p{4cm}} % Reduced to 3 columns
\toprule
\textbf{Process} & \textbf{Text} & \textbf{Source Title (Link)} \\ \midrule

\textit{Script Filtering} & ᎬᎾ ᏑᎵ ᎧᏃᎮᏓ ᎦᏂᏓᏛ,ᎠᏖᎳ ᎤᏁᎦ [lining that extends from the tip of its wing along the rear of the wing to its body, ᏃᎴ a head with reddish skin. The ᎬᏂᎨ ᏑᎵ has a shorter tail, silver splotches near the tips of its wings, ᏃᎴ a head with grayish skin. Both ᎬᏂᎨ ᏑᎵ ᏃᎴ ᎬᎾ ᏑᎵ ᎯᎸᏍᎩ ᏥᏍᏆ locate the dead animals or carrion they feed upon by both sight ᏃᎴ smell. & \href{https://chr.wikipedia.org/wiki/\%E1\%8F\%A5\%E1\%8F\%8D\%E1\%8F\%86}{ᏥᏍᏆ} \\ \addlinespace

& ᎠᏂᏣᎳᎩ ᎠᏂᏬᏂᏍᎩ polysynthetic ᎦᏬᏂᎯᏍᏗ. ᏣᎳᎩ ᎦᏬᏂᎯᏍᏗ ᎤᏐᏱ ᏏᏓᏁᎳ Iroquoian ᏧᏂᏬᏂᎯᏍᏗ, ᎠᏎᏃ, ᏍᏈᏍᏓ ᏗᏑᏕᏘᏴᏓ ᏥᎨᏒ, ᏐᏉ ᎦᏬᏂᎯᏍᏗ ᏱᎩ. ᎠᏂᏣᎳᎩ ᎠᏂᏴᏫᏯ ᏧᏃᎯᏳ, ᎤᏁᎳᏅᎯ ᎦᏬᏂᎯᏍᏗ ᎤᏮᏏ ᏥᎨᏒᎢ. 1820Ꮫ, ᏍᏏᏉᏯ, ᎠᎴ George Gist, ᎯᎠ ᏣᎳᎩ ᏗᎪᏪᎵ ᎪᏪᎵᏍᎨᎢ. ᏃᏉ ᎯᎠ Unicode ᎢᎩᎭ (ᎭᏫᎾᏗᏢ). ᎯᎠᏃ ᎦᏓᎧ, ᎡᎵᏉ, ᏣᎳᎩ ᎢᎪᏪᎶᏗ ᎤᏅᏥᏓ ᎤᎾᎦᎵᏍᎩ. ᎯᎠᏍᏊ ᏄᏓᎴᎢ ᏮᏎ ᎤᏰᎵᏛ ᏂᎦᎵᏍᏗᎮᎢ: 2007 ᏥᎨᏒ, ᎥᎭᏴᎵ ᏗᏰᎴᎩ (Myrtle Driver Johnson) ᎠᎬᏱ ᎢᏳᏩᎧᏔ 1800Ꮝ ᏥᎨᏒ, Charles Frazier ᏗᎪᏪᎵ ᎤᏤᎵ, ᏦᎦᏚ ᏅᏙ: ᏥᎨᎬᏬᎥᏗᏍᎨᎢ ᏣᎳᎩ ᎠᏁᏍᏗᏍᎩ ᎪᏪᎵᏍᎨᎢ. & \href{https://chr.wikipedia.org/wiki/\%E1\%8F\%A3\%E1\%8E\%B3\%E1\%8E\%A9\_\%E1\%8E\%A7\%E1\%8F\%AC\%E1\%8F\%82\%E1\%8E\%AF\%E1\%8F\%8D\%E1\%8F\%97}{ᏣᎳᎩ ᎧᏬᏂᎯᏍᏗ} \\ \addlinespace

& ᎹᎶᎪ (Ma-lo-go), ᏑᏓᎴᎩ ᎤᎾᏓᏟᏌᎲ ᎤᎬᏫᏳᎯ ᎹᎶᎪ (Al-Mamlakah al-Maɣribiyya), ᎠᏰᎵ ᎩᏄᏓᏕᎩ ᎠᏣᏱ. & \href{https://chr.wikipedia.org/wiki/\%E1\%8E\%A4\%E1\%8E\%AC\%E1\%8F\%AB\%E1\%8F\%B3\%E1\%8E\%AF}{ᎤᎬᏫᏳᎯ} \\ \midrule

\textit{Exact Match} & Tiŋ ŋɔ nyɛla tiŋ' shɛli din be Zabzugu Municipal Assembly, Northern Region, Ghana tiŋ gbaŋ ni. & \href{https://dag.wikipedia.org/wiki/Sawieb}{Sawieb} \\ \addlinespace
& Tiŋ ŋɔ nyɛla tiŋ' shɛli din be Zabzugu Municipal Assembly, Northern Region, Ghana tiŋ gbaŋ ni. & \href{https://dag.wikipedia.org/wiki/Tayondo}{Tayondo} \\ \addlinespace
& Tiŋ ŋɔ nyɛla tiŋ' shɛli din be Zabzugu Municipal Assembly, Northern Region, Ghana tiŋ gbaŋ ni. & \href{https://dag.wikipedia.org/wiki/Poagmado\_Polinaayili}{Poagmado Polinaayili} \\ \midrule

\textit{MinHash} & Say Central Visayas et rehiyon na Filipinas. Unong ed 1 Hulyo 2024 census, say populasyon to et 6,640,875 totoo tan 1,966,588 abong. Walay kabaleg tan sukat to ya 15,895.66 sq. km. & \href{https://pag.wikipedia.org/wiki/Central\_Visayas}{Central Visayas} \\ \addlinespace
& Say Ilocos Region et rehiyon na Filipinas. Unong ed 1 Hulyo 2024 census, say populasyon to et 5,342,453 totoo tan 1,306,256 abong. Walay kabaleg tan sukat to ya 13,012.60 sq. km. & \href{https://pag.wikipedia.org/wiki/Ilocos\_Region}{Ilocos Region} \\ \addlinespace
& Say Central Luzon et rehiyon na Filipinas. Unong ed 1 Hulyo 2024 census, say populasyon to et 12,989,074 totoo tan 2,511,783 abong. Walay kabaleg tan sukat to ya 22,014.63 sq. km. & \href{https://pag.wikipedia.org/wiki/Central\_Luzon}{Central Luzon} \\
\bottomrule
\end{tabularx}
    \caption{\footnotesize Examples of Wikipedia articles targeted by each filtering process. The selected languages are: Cherokee (\texttt{chr}, top), Dagbani (\texttt{dag}, middle), and Pangasinan (\texttt{pag}, bottom).}
    \label{tab:process-filtering}
\end{table*}

\section{Is Wikipedia `High Quality'?}\label{sec:prefiltering}

\citet{albalak2024surveydataselectionlanguage} outline a set of best practices for quality data selection and filtering. In this work, we examine the effects of their proposed model-agnostic filtering strategies,\footnote{We exclude classifier-based and perplexity-based filtering strategies measuring quality relative to a selection of data.} as applied to Wikipedia. Conceptually, we can recognize such filters as belonging to two distinct categories based on their underlying objective: 1) procedures for removing undesirable material from Wikipedia, such as foreign scripts and duplicate articles; and 2) heuristics that aim to enhance the existing data distribution by removing otherwise lower quality text according to a given metric (e.g. article length in characters).

We apply select filters from the first category to identify undesirable content in non-English Wikipedias --- namely script filtering and deduplication. We posit that if these filters remove a substantial amount of text from a given Wikipedia dump, that content was not of high quality to begin with. Ultimately, we find that 11.68\% and 29.36\% of total characters and documents are removed from the raw Wikipedia, respectively, as shown in Figure \ref{fig:main} (top). In what follows, we will analyse each filtering step, as well as the languages most affected by them. 

\paragraph{Script Filtering} Applying an existing language identification model to Wikipedia is a challenge due to data contamination -- most models supporting all of Wikipedia's languages have generally also been trained on Wikipedia 
\citep[see e.g.][]{kargaran-etal-2023-glotlid}. Thus, we employ script-based filtering as a proxy for language identification by filtering characters in the Wikipedia that are not in the officially documented script(s)\footnote{For example, the Konkani Wikipedia features 3 scripts -- Devanagari, Latin and Kannada.} for the concerned language (see Appendix \ref{app:script} for more details).\footnote{\url{https://en.wikipedia.org/wiki/List_of_Wikipedias}} We find that 3.33\% of characters in non-English Wikipedias are removed by this process.\footnote{We acknowledge that this methodology would not be effective for English articles in non-English Wikipedias written in the Latin script. However, this
would require a language ID model that has not been trained on Wikipedia. We also attempted to use a language ID model to exclusively classify and filter English text, but we found that this led to over-filtering and misclassification for some Creoles and Pidgins, eg. Naija.} Interestingly, the relative amount of removed documents is much smaller at 0.01\% (7782 total), which indicates that this text stems from very long, potentially copied and pasted articles. For example, consider the \href{https://web.archive.org/web/20251230045620/https://as.wikipedia.org/wiki/%E0%A6%86%E0%A6%B2%E0%A7%87%E0%A6%95%E0%A6%9C%E0%A7%87%E0%A6%A3%E0%A7%8D%E0%A6%A1%E0%A6%BE%E0%A7%B0_%E0%A6%AB%E0%A7%8D%E0%A6%B2%E0%A7%87%E0%A6%AE%E0%A6%BF%E0%A6%82}{article for Alexander Fleming in Assamese}, which is mostly a direct copy of the corresponding English version. This particular article is a symptom of the problems associated with the use of machine translation tools\footnote{e.g. \url{https://www.mediawiki.org/wiki/Content_translation}} in Wikipedia content creation, which can lead to entire sections of articles being mistranslated, or not translated at all \cite{wilsonWikipediaHasGoogle2019a}. With this in mind, script filtering can be utilized as a crude way to remove mistranslated content. However, the problem of detecting translationese in Wikipedia is complex and well documented \cite{pageHowAIWikipedia} --- especially for low-resource languages --- and is not easy to mitigate without human evaluation.

Outside of translation artefacts, it is important to note that the vast majority of characters removed from script filtering stem from low-resource or extinct languages (for an example, refer to Table \ref{tab:process-filtering} (top)), such as Cree (\texttt{cr}, 187 documents), Cherokee (\texttt{chr}, 1113), or Gothic (\texttt{got}, 1013). In most cases, these are Latin characters corresponding to placeholder information, transliterated named entities or embedded links (if improperly formatted). In many cases, Wikipedia is among the few public resources that collects text for such languages, making it crucial to apply filters that preserve authentic, language-specific content --- especially before downstream use in dataset creation, tokenizer training, or other applications. 

\paragraph{Exact-match Deduplication} We employ a hashing function to identify and remove articles which have the exact same text, but a different unique ID or title. This removes 0.16\% and 1\% of all characters and documents, respectively. However, we find that some Wikipedias are disproportionately affected by this process. For example, the Yorùbá Wikipedia (\texttt{yo}, 33,819 documents) loses almost 60\% of its documents --- many of which contain a single token \textit{\`{I}t\d{o}\'{k}as\'{\i}} (Reference). In general, articles typically removed by exact-match deduplication are placeholders containing filler or templatic text and generally contain little usable information. As an illustration of this, refer to Table \ref{tab:process-filtering} (middle), which shows a template assigned to 48 distinct municipality pages in the Dagbani (\texttt{dag}) Wikipedia. Accounting for such phenomena --- especially at the document level --- ensures that resource statistics are not artificially inflated and accurately represent the amount of authentic data available in a given language. 

% Likewise, the Dagbani Wikipedia (\texttt{dag}, 10,071) features \href{https://web.archive.org/web/20260112204631/https://dag.wikipedia.org/wiki/MOS:NEOPRONOUN}{a single, moderately long article} that spans 44,945 characters and is repeated across 80 unique pages, amounting to 6.52\% of its total character count. 

\paragraph{MinHash Deduplication} While exact-match is able to filter identical documents, it cannot account for partial duplicates --- \textit{i.e.}, sets of documents containing mostly identical text, but varying in terms of single sentences, paragraphs, or even hyperlinks. For example, the Yorùbá articles for the \href{https://web.archive.org/web/20260108213956/https://yo.wikipedia.org/wiki/B%C3%A0l%C3%BA%E1%B9%A3%C3%AC}{Baluch people} and the \href{https://web.archive.org/web/20260101194428/https://yo.wikipedia.org/wiki/%C3%88d%C3%A8_B%C3%A0l%C3%B3%E1%B9%A3%C3%AC}{Baluchi language} are largely identical, with the latter containing an additional language data panel (see Table \ref{tab:process-filtering} (bottom) for another example in Pangasinan). To capture such cases, we perform \textit{MinHash deduplication} with locality sensitive hashing (LSH) and discount any article that returns a Jaccard similarity over 0.85. We find that a staggering 28.33\% of all non-English Wikipedia articles (and 8.18\% characters) are affected by this procedure, largely consisting of Wikipedias known to be primarily bot generated (more in \S\ref{sec:bot}). Here, it is important to note that many entries with high MinHash removal rates are likewise among the largest in terms of total document count prior to filtering. These include Cebuano (\texttt{ceb}, 6.1 million documents, 80\% removed), Swedish (\texttt{sv}, 2.6, 47\%)\footnote{see \citet{WhyAreThere}}, Dutch (\texttt{nl}, 2.1, 35\%), and Vietnamese (\texttt{vi}, 1.3, 58\%) --- the raw data of which was ostensibly used to train the highly influential multilingual BERT model.\footnote{\url{https://github.com/google-research/bert/blob/master/multilingual.md}}

\section{Characterizing Quality}\label{sec:tiers}

\begin{figure*}[htbp]
    \centering
    \includegraphics[width=0.55\linewidth]{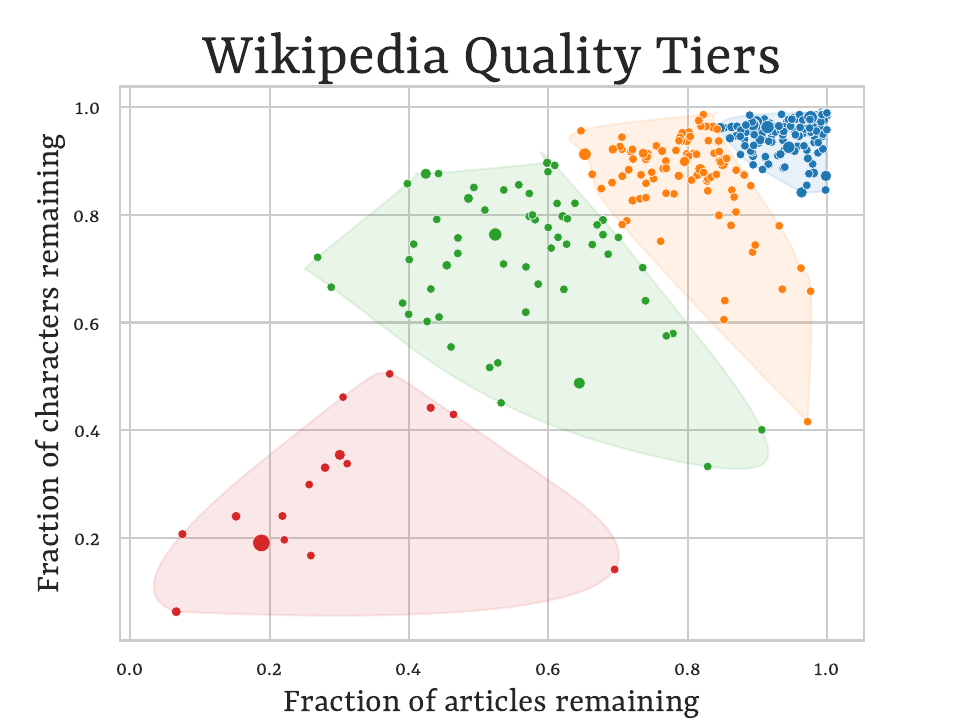}
    \includegraphics[width=0.43\linewidth]{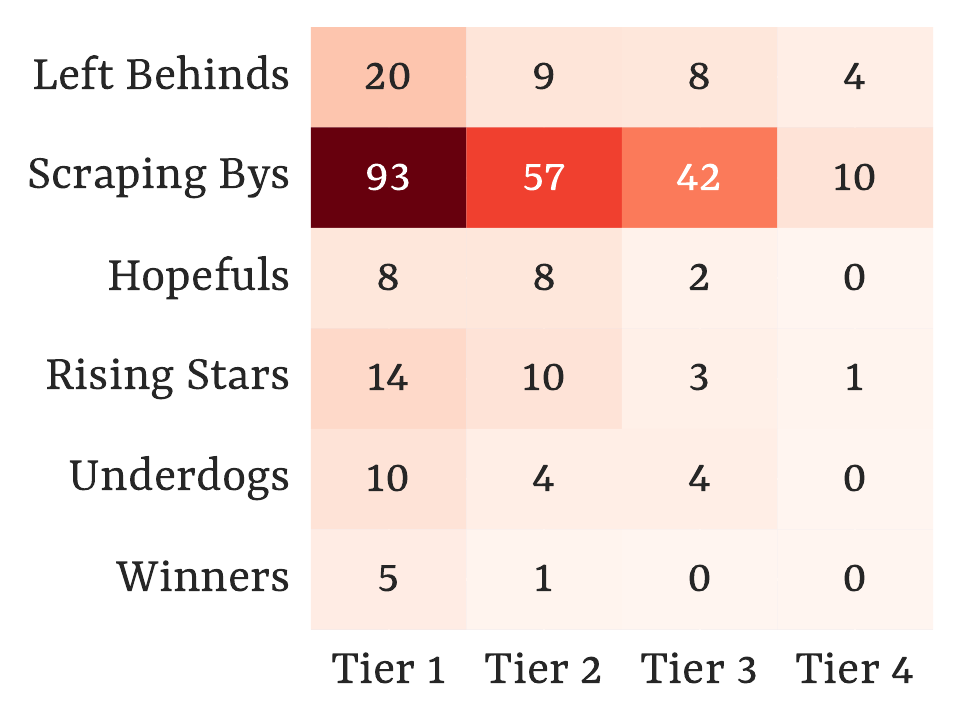}
    \caption{\footnotesize \textit{Left}: Quality clusters based on the fraction of documents and characters removed after filtering, respectively. Clusters correspond to {\color{blue} \textsc{Tier 1}} (blue), {\color{orange} \textsc{Tier 2}} (orange),  {\color{green} \textsc{Tier 3}} (green) and {\color{red} \textsc{Tier 4}} (red). Each point is weighed by each Wikipedia's unfiltered document count. \textit{Right}: A confusion matrix comparing the \citet{joshi-etal-2020-state} and the Wikipedia quality tiers.}
    \label{fig:wiki-cluster}
\end{figure*}

Having demonstrated the effect of filtering, we now attempt to shed light on how our approach relates to a generalized notion of dataset quality (as it pertains to Wikipedia). To do so, we first categorize all non-English Wikipedias into 4 tiers --- \textsc{Tier 1, 2, 3} and \textsc{4} --- based on how much data is removed during filtering (ranked from least to most). We perform $k$-means clustering, where each Wikipedia is represented by the percentage of remaining documents and characters. Based on several iterations of the clustering algorithm and manual inspections, we arrive at the ranking illustrated in Figure \ref{fig:wiki-cluster} (left). A list of all Wikipedias and their respective quality tiers can be found in Table \ref{tab:tiers}. 

\subsection{Resource Availability}

At first glance, our quality ranking might seem to categorize Wikipedias based on resource availability. To assess how accurate this impression is, we compare our classification with the taxonomy proposed by \citet{joshi-etal-2020-state}, which divides the world's languages into 6 categories according to the volume of labelled and unlabelled data available for them (with the latter measured by Wikipedia document counts). Figure \ref{fig:wiki-cluster} (right) presents a comparison between their taxonomy and our quality tiers. As expected, we observe that \citet{joshi-etal-2020-state}'s categorization has a strong correlation with Wikipedia size, but demonstrates weak alignment with our quality taxonomy, suggesting that quality is not simply a direct consequence of resource availability.

While we observe that most \textsc{Winners} unsurprisingly are allocated to \textsc{Tier 1}, Arabic falls to \textsc{Tier 2}. Upon a cursory inspection, we find that 15\% of the original Arabic Wikipedia is removed via MinHash deduplication, indicating that it contains a sizeable volume of potentially templatic or bot-generated text. We notice a similar effect for other higher-resourced \textsc{Underdogs} and \textsc{Rising Stars} languages, which are placed in \textsc{Tier 3} (Swedish, Vietnamese, Serbian and Basque) and \textsc{Tier 4} (Cebuano) --- also due to their vulnerability to MinHash (as noted in \S \ref{sec:prefiltering}). Going further, we find that 113 of the lowest resourced languages classified as \textsc{Scraping Bys} and \textsc{Left Behinds} (35\% of Wikipedia in total) are likewise categorized as having \textsc{Tier 1} Wikipedias, implying the involvement of dedicated moderation communities. Conversely, we also observe many \textsc{Scraping Bys} and \textsc{Left Behinds} languages falling into the \textsc{Tier 3} and \textsc{Tier 4} categories, such as Yorùbá and Twi, indicating that they contain serious quality issues that can be harmful to low-resource NLP.

\subsection{Comparison to \textsc{Depth+}}
In its official statistics, Wikimedia provides a \textsc{Depth} metric (alternatively, \textit{editing depth}), designed to serve as a proxy measure for a given Wikipedia's `collaborative quality'.\footnote{\url{https://meta.wikimedia.org/wiki/Wikipedia_article_depth}} However, \citet{alshahrani-etal-2023-depth} caution against utilizing it for such purposes. Specifically, they discuss the metric's sensitivity to bot content (articles and edits) and excessive editing (`edit wars'), as well as its inability to consolidate broader user activity and account for lower-resourced Wikipedias. As such, they propose a modified metric, \textsc{Depth+}:

\begin{equation}
    \mathrm{Depth}^{+} = \mathrm{Editors} \cdot \frac{\mathrm{Edits}}{\mathrm{Total}} \cdot \frac{\mathrm{Articles}}{\mathrm{Non-Articles}}
\end{equation}

\citet{alshahrani-etal-2023-depth} show that \textsc{Depth+} is able to adequately account for bot-generated content, correctly de-ranking Wikipedias such as Vietnamese, Serbo-Croatian, and Cebuano. Conversely, they demonstrate that \textsc{Depth+} likewise includes several low-resource Wikipedias, such as Cree, Tigre, and Bangla towards the top of the ranking, highlighting their users' supposedly high degree of collaboration. 

We find a significant positive correlation between our filtering procedure (measured as the ratio of percentage of documents and characters retained) with \textsc{Depth+} ($\rho=0.31, p<0.001$). Thus, both metrics appear to capture similar notions of quality. However, approximately half of all Wikipedias yield \textsc{Depth+} values of less than 0.3, and a majority very close to 0. The rest yield significantly higher \textsc{Depth+} values (excluding English, German is highest with 36.04) resulting in many outliers that skew the distribution. We surmise that this is because the number of edits is disparately different across Wikipedia, effectively privileging Wikipedias where active users comprise a larger percentage of total users --- a proxy for `collaborative quality'. However, this does not shed much insight on the quality of individual articles written by one-time, potentially domain-expert authors. 

\begin{figure*}[ht!]
    \centering
    \includegraphics[width=\linewidth]{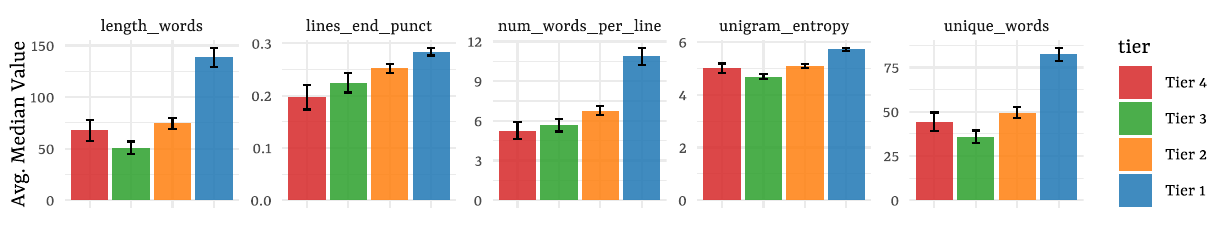}
    \caption{\footnotesize Aggregate median heuristic values by Tier. Spread across Wikipedias indicated by error bars.}
    \label{fig:heuristics}
\end{figure*}

\subsection{Bot generation} \label{sec:bot}
We have observed that our filters remove a sizeable volume of articles from Wikipedias known to contain proliferate bot-generated content.
To better understand if we are indeed filtering out such content, we use \citet{alshahrani-etal-2023-depth}'s estimation of bot-generated articles across Wikipedia, which draws from user registration information and other metadata. Indeed, we find that there is a strong negative correlation ($\rho=-0.56, p<0.001$) between the percentage of bot articles in a given Wikipedia and the percentage of articles remaining after filtering. We consider this to be an encouraging finding, provided that our data filters are simple, pre-defined processes that state obvious assumptions about our ideal data distribution. Additionally, we do not rely on specialized bot-detection models trained on other distributions. 

We further hypothesize that MinHash deduplication plays a large role in removing bot articles, given its ability to fuzzy-match strings across documents. Here, we observe that the percentage of documents filtered out by MinHash yields an even higher correlation with bot article ratio ($\rho=0.63, p<0.001$), indicating that this process is an adequate proxy for bot article removal in Wikipedia. Unsurprisingly, the Wikipedias with the highest percentage of bot content according to \citet{alshahrani-etal-2023-depth} --- Cebuano (99\%), Waray (90\%), Swedish (68\%) --- tend to have much (if not most) of their content removed by MinHash. 

\subsection{Heuristics as Quality Measures}\label{sec:heuristics} 
\citet{albalak2024surveydataselectionlanguage} define \textit{heuristic filtering} as a process that computes document-level statistics to eliminate low-quality data that falls below a specified threshold. Heuristics, in this sense, refer to simple metrics that aim to numerically characterise document content --- for example, the number of characters contained in a given document. Though we do not attempt to filter Wikipedia according to these metrics (see Appendix \ref{app:data}), we are nonetheless interested in how their distributions align with our quality tiers. In order to investigate this, we calculate five common and weakly correlated heuristics, as implemented in the Red Pajama \citep{together2023redpajama} dataset cleaning pipeline: length in words, number of unique words, unigram entropy, average number of words per line, and fraction of lines ending in punctuation.\footnote{The remaining heuristics are largely tailored to web-text and are thus not directly relevant to Wikipedia, such as mentions of “javascript” or “lorem ipsum”, or excessive within-document string duplication.} Due to the fact that many heuristics tend to follow an exponential distribution, we calculate each metric at the document level and record the median value for each Wikipedia (refer to Appendix \ref{app:heuristics_per_tier} for each median heuristic value per edition, per tier).  

% \footnote{Filtering Wikipedia using heuristics requires an arbitrary threshold to serve as a cut-off between `good' and `bad' quality. Existing thresholds (e.g. Gopher \cite{rae2021scaling}) are normally tuned for English, and not directly applicable to other languages. Though we attempted to filter Wikipedia articles on thresholds defined by language-specific data distributions (the procedure is described in \S \ref{app:data}) and conducted similar downstream experiments as detailed in \S \ref{sec:downstream}, we found little gains over simple script filtering and deduplication. We defer language-specific heuristic discovery for future work.}

In averaging median heuristics across Wikipedias, we observe a clear, monotonic pattern that neatly matches our ranking: \textsc{Tier 1} produces the best values (in terms of polarity), followed by \textsc{Tier 2} and then \textsc{Tier 3} (Figure \ref{fig:heuristics}). This is largely intuitive: high quality articles can be assumed to be long (\texttt{length\_words}), verbose (\texttt{unique\_words}), less lexically predictable (\texttt{unigram\_entropy}), as well as to contain longer (\texttt{num\_words\_per\_line}) and better-formatted paragraphs (\texttt{lines\_end\_punct}). The only consistent exception is that of \textsc{Tier 4}, which does not follow monotonicity for 3 out of 5 heuristics, landing in between Tiers \textsc{2} and \textsc{4}. However, this is largely expected: \textsc{Tier 4} contains Wikipedias that are especially vulnerable to select filtering steps, which can affect heuristic measurement in different ways. For example, Cebuano scores high on \texttt{lines\_end\_punct} due to being largely bot generated, but likewise yields low \texttt{unigram\_entropy} values for the same reason. Indeed, we can confirm this to be the case by observing the spread of median heuristic values, with \textsc{Tier 4} being the highest for each metric. 

\begin{figure*}[ht!]
    \centering
    \includegraphics[width=0.3\linewidth]{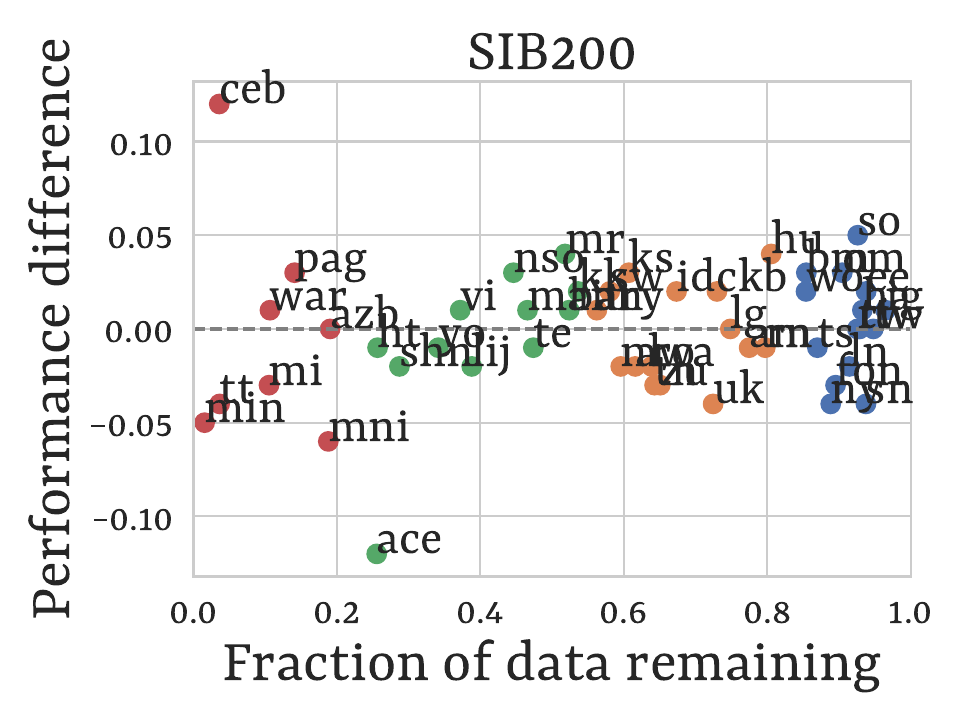}
    \includegraphics[width=0.3\linewidth]{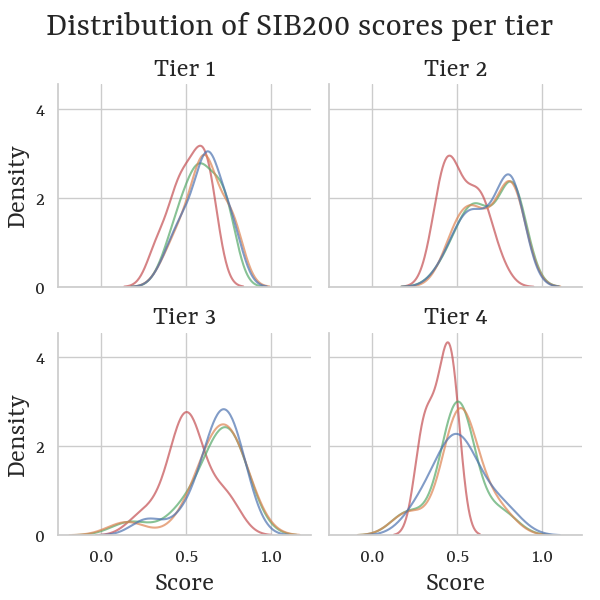}
    \includegraphics[width=0.3\linewidth]{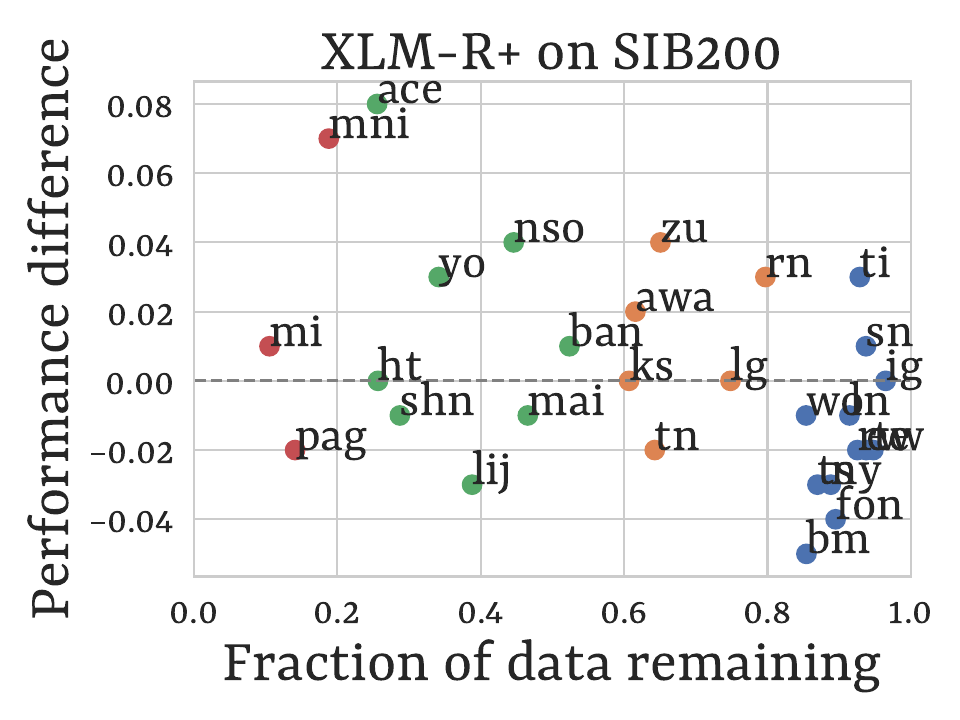}
    \caption{\footnotesize \textit{Left}: Performance difference over \textsc{SIB200} after monolingual pretraining on \expRaw and \expPrimary data; positive values indicate preference for the latter. Colours correspond to quality tiers described in \S \ref{sec:tiers} \textit{Center}: KDE plots of \textsc{SIB200} score distributions of \expRaw, \expPrimary, and \expRaw models, organized by tier. \textit{Right}: Performance difference over \textsc{SIB200} after language adaptation. All scores are an average of 5 runs. Full result tables for all tasks are in Appendix \ref{app:hypers}.}
    \label{fig:sib200}
\end{figure*}

% \vspace{-1cm}
\section{Downstream Evaluation}\label{sec:downstream}
The most salient use case for Wikipedia within NLP has been language modelling. This section examines how low-quality Wikipedia content affects this domain across three different contexts: training models in a single language, adapting models to new languages, and training multilingual models. This analysis compares two datasets: the complete, unfiltered Wikipedia data (\expRaw) versus data cleaned using the methods outlined in \S \ref{sec:prefiltering} (\expPrimary).\footnote{Full training details for all 3 setups are in Appendix \ref{app:hypers}.} The underlying hypothesis is straightforward --- if the performance difference ($\Delta$) between models trained on \expPrimary and \expRaw is negligible, this would confirm that the removed data was indeed low quality.

\subsection{Monolingual Pretraining} \label{sec:pretraining}
In order to minimize confounding factors such as language imbalance or poorly calibrated tokenizers, we pretrain mini monolingual DeBERTa \cite{he2021deberta} models (approximately 10 million parameters) on \expRaw and \expPrimary data from 50 languages and evaluate them on topic classification with \textsc{SIB200} \cite{adelani-etal-2024-sib}.\footnote{We also evaluate these models on MasakhaNER, AfriSenti-Twitter and MasakhaNEWS. However, since the results for these tasks are similar to \textsc{SIB200}, we include them in Appendix \ref{app:hypers} in the interest of brevity.} 

\begin{figure*}[t]
    \centering
    \includegraphics[width=0.3\linewidth]{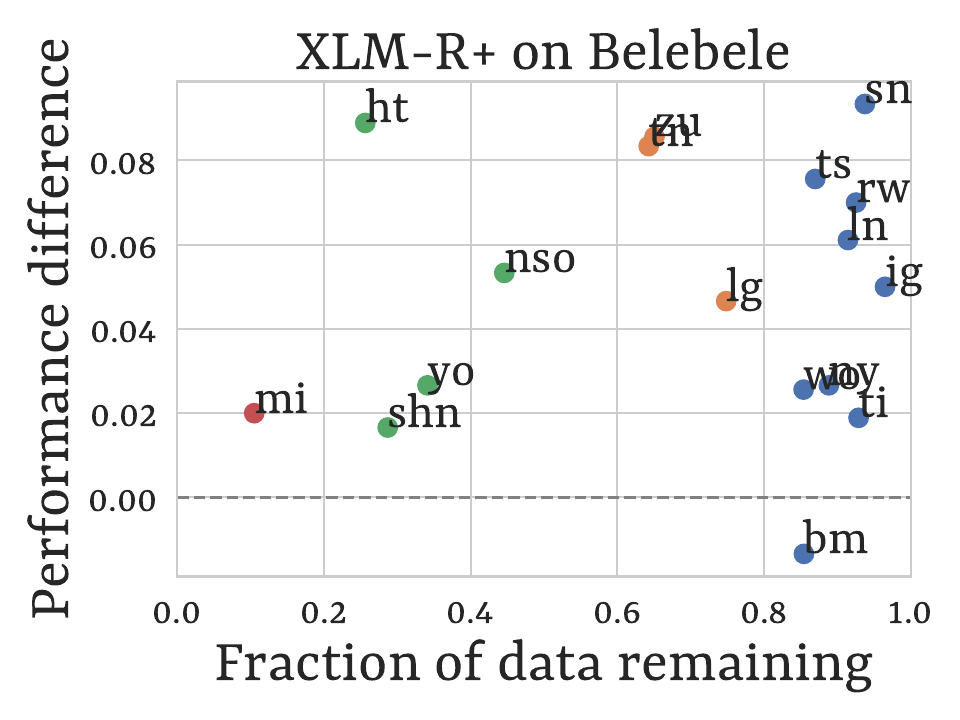}
    \includegraphics[width=0.3\linewidth]{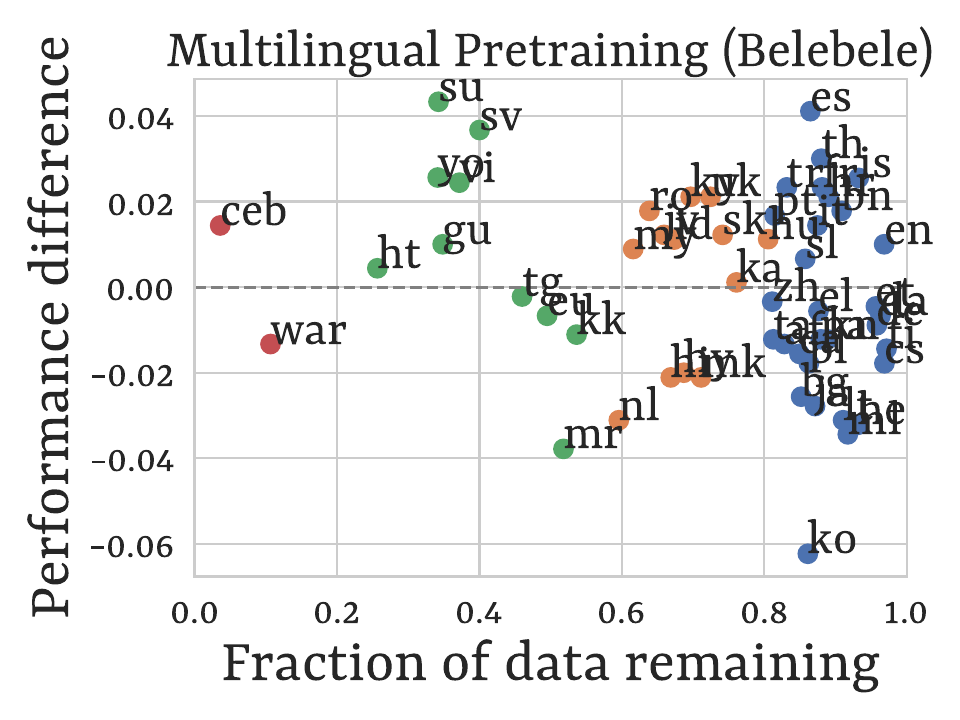}
    \includegraphics[width=0.3\linewidth]{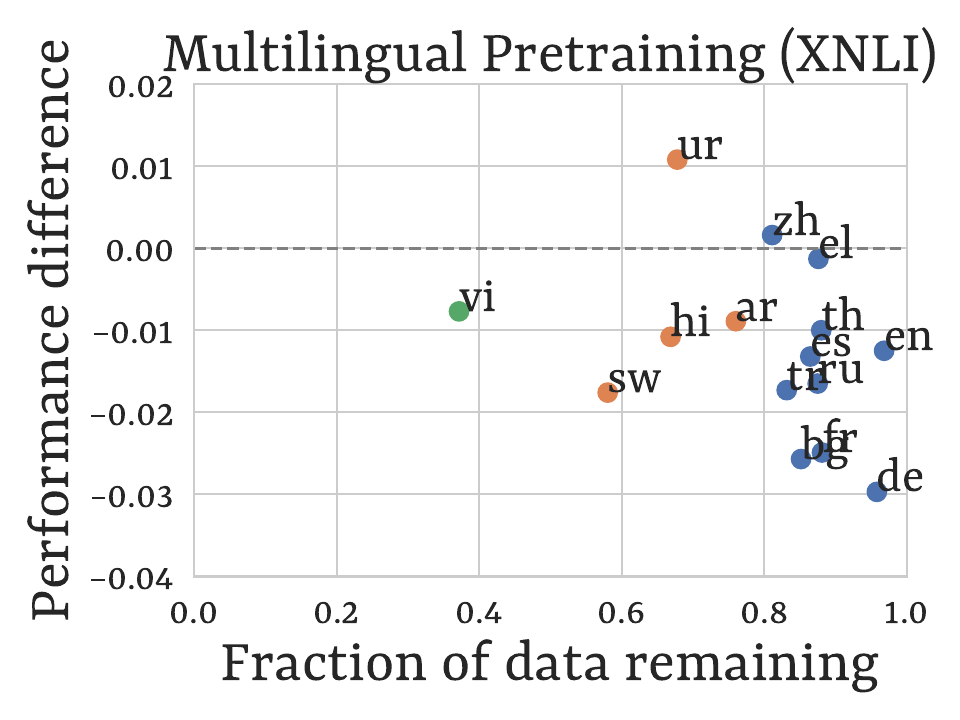}
    \caption{\footnotesize \textit{Left}: Performance difference over \textsc{Belebele} after language adaptation. \textit{Center}: Performance difference over \textsc{Belebele} after multilingual pretraining. \textit{Right}: Performance difference over \textsc{XNLI} after multilingual pretraining.}
    \label{fig:adapt-and-multi}
\end{figure*}

The left panel of Figure \ref{fig:sib200} shows $\Delta$ for each Wikipedia edition relative to the proportion of retained data. Here, we observe that most values fall within the $(-0.05, 0.05)$ range, suggesting that the removed content is indeed low quality and does not contribute meaningfully to model training. A notable example is Cebuano (\texttt{ceb}), which achieves a 12\% performance improvement with \expPrimary — a gain we largely attribute to the deduplication process eliminating most bot-generated content. In contrast, Acehnese (\texttt{ace}) appears negatively impacted by filtering, with $\Delta = -0.12$. Closer examination reveals that while this Wikipedia is similarly affected by MinHash deduplication (losing 42\% of characters and 54\% of documents), it contains far fewer raw documents than Cebuano --- only 13,000 compared to 6 million. This suggests that such substantial data reduction can hurt performance in low-resource scenarios, even when the removed content resembles bot-generated content.

In order to confirm that \expPrimary indeed removes low quality documents, we conduct an additional control experiment (\expRandom), wherein we sample $n$ random articles from \expRaw to match the size of \expPrimary. Figure \ref{fig:sib200} (center) shows the distribution of scores for all three conditions separated by tier. Here, we observe that the distribution of the \expRandom scores is, on average, substantially lower than the other methods, with more pronounced differences seen for \textsc{Tiers 2, 3} and \textsc{4}. This is notable, since \expPrimary removes a larger proportion of data for lower-ranked Wikipedias than those in \textsc{Tier 1}.

\subsection{Language Adaptation}
As a more common use-case for low-resource languages \cite{pfeiffer-etal-2020-mad, pfeiffer-etal-2022-lifting}, we adapt XLM-R \citep{conneau2020unsupervised} to Wikipedias that were not included in its training data, according to the experimental protocol of \citet{alabi-etal-2022-adapting}. 
% We use the original XLM-R tokeniser and further pretrain it on individual Wikipedias across \expRaw and \expPrimary settings.
Following this, we fine-tune each model on \textsc{SIB200} and \textsc{Belebele} \cite{bandarkar-etal-2024-belebele} --- a cross-lingual reading comprehension benchmark.\footnote{We fine-tune models on English data --- as recommended in the \textsc{Belebele} paper --- and evaluate them on the target languages in a zero shot manner. This setup is not possible with the monolingual models due to tokeniser restrictions.}

The right panel of Figure \ref{fig:sib200} shows that \textsc{Tier 1} $\Delta$ values remain within the $(-0.05, 0.05)$ range, mirroring the monolingual pretraining results. However, a different pattern emerges for the remaining tiers: most exhibit performance improvements after adapting XLM-R with \expPrimary data. Acehnese is again a notable highlight here, returning $\Delta = 0.08$ --- an indication that a well-calibrated model can benefit from smaller quantities of higher-quality data. We observe even better results on \textsc{Belebele} (Figure \ref{fig:adapt-and-multi}, left), with all but one model performing better in the \expPrimary setting. Previous work has shown that quality of pretraining data is more impactful to performance in a zero-shot setting than when fine-tuning the model on the target language \citep{tatariya-etal-2023-transfer}. These findings provide additional evidence that the filtered content was indeed low quality.
% \vspace{-1cm}

\subsection{Multilingual Pretraining}
In our third experiment, we mimic the setup of multilingual BERT \cite{devlin-etal-2019-bert} and pretrain two base-size multilingual models (approximately 100 million parameters) on \expRaw and \expPrimary data for the top 100 Wikipedias sorted by raw document count. We evaluate these models on \textsc{Belebele} and \textsc{XNLI} \citep{conneau-etal-2018-xnli}, after performing English task-specific fine-tuning. Regarding the former dataset, we observe similar trends to the previous two experiments: $\Delta$ values are largely within the $(-0.05, 0.05)$ range, with more lower-tier Wikipedias appearing to benefit from filtering than \textsc{Tier 1} (Figure \ref{fig:adapt-and-multi}, center). However, we do not observe the same effect as seen for XLM-R (Figure \ref{fig:adapt-and-multi}, left), with all but two models dropping in performance (Figure \ref{fig:adapt-and-multi}, right) --- albeit very slightly and within a very narrow range of $(-0.03, 0.01)$.

\section{Conclusion}

In this study, we demonstrated that Wikipedia data quality varies substantially across language editions, challenging the assumption that it represents a consistent and impeccably curated resource. In applying a set of common data cleaning techniques to the entire set of non-English editions, we found that approximately 30\% of documents and 12\% of characters thereof represent low-quality content --- including varying degrees of foreign-script contamination, templates repeated across hundreds of URLs, and a proliferation of bot-generated articles. We consolidated these results into a tiered ranking of all non-English Wikipedia editions, which showed strong correspondence with alternative notions of dataset quality, such as \textsc{Depth+}, proportion of bot-generated articles, and other heuristic measures such as document length and unigram entropy. Lastly, in a downstream evaluation across three practical applications --- monolingual pretraining, language adaptation, and multilingual pretraining --- we confirmed that models trained on a filtered Wikipedia generally match or exceed the performance of those trained on raw data, with lower quality editions generally benefitting more. 

Ultimately, the findings reported here serve to underscore the necessity of rigorous data auditing --- even when working with a trusted data source like Wikipedia. While our approach is highly automated, it nonetheless demonstrates that simple, model-agnostic preprocessing can go a long way in discerning between good and bad quality data. As Wikipedia continues to serve and evolve as a foundational resource for NLP, understanding and accounting for these quality variations becomes essential for developing more equitable and effective language technologies. This is particularly true for low-resource languages, for which only several thousand (or even hundred) articles are attested. As such, we believe that future work should aim to engage with Wikipedia editing communities to better understand the sources of quality variation and identify language-specific considerations that automated filtering may overlook.

\section{Limitations}
\paragraph{Use of Encoder Models} Our experimental setup has been limited to encoder-only models since already existing decoder-only models such as Llama \cite{touvron2023llama2openfoundation} and Deepseek \cite{deepseekai2025deepseekv3technicalreport} have already most likely been trained on Wikipedia. We attempted similar language adaptation experiments for decoder models, and in our experiments with Deepseek, for instance, we immediately observed low losses within the first few 100 steps of continued pretraining, indicating that the model had already been trained on Wikipedia. We also found that these models do not perform well especially for low resource languages such as the ones included in the paper \cite{adelani-etal-2025-irokobench}, with no improvement even after continued pretraining --- likely due to the comparatively small size of individual Wikipedias \cite{zhang2024snakmodellessonslearnedtraining}. Moreover, benchmarks for decoder models rely on English prompts to evaluate multilingual language performance, which has been noted as a problematic paradigm (see, e.g. \citet{poelman-lhoneux-2025-roles}), and which also restricts us from training and evaluating monolingual decoder-only models like \citet{chang2024goldfishmonolinguallanguagemodels}. 

\paragraph{Evaluation Constraints} Our evaluation has been restricted to benchmarks that have not been created from Wikipedia in order to not introduce additional confounds across the pretraining and fine-tuning stages. Moreover, we note that most multilingual evaluation datasets are designed to evaluate multilingual models in zero-shot or few-shot settings, without in-language train sets (like \textsc{Belebele} and \textsc{XNLI}), making them unusable to evaluate monolingual models due to tokenization restrictions (e.g. a tokenizer trained on English text cannot transfer to Japanese by design, as both languages employ different alphabets). Our evaluation setup is, as a consequence, limited. 

\section{Acknowledgements}
For KT, AK, WP and MDL the computational resources and services used were provided by the VSC (Flemish Supercomputer Center), funded by the Research Foundation - Flanders (FWO) and the Flemish Government - department EWI. AK was partially funded by a Google Research Award.
WP is funded by a KU Leuven Bijzonder Onderzoeksfonds C1 project with reference C14/23/096. EP, HL and JB are funded by the Carlsberg Foundation, under the Semper Ardens: Accelerate programme (project nr. CF210454).
MB is supported by TrustLLM funded by Horizon Europe GA~101135671, with computational resources provided by the National Academic Infrastructure for Supercomputing in Sweden (NAISS) partially funded by the Swedish Research Council through grant agreement no.\ 2024/22-745.
We would also like to thank Isaac Caswell, Colin Cherry, and Markus Freitag at Google Translate Research for their insightful reviews on the initial draft.

\bibliography{new}
% \bibliographystyle{acl_natbib}

% \onecolumn
% \newpage
\appendix

\section{Script Filtering}\label{app:script}
For script filtering, we utilize a regular expression that matches unicode characters belonging to the Wikipedia-documented ISO-15924 script range (e.g. \texttt{Latn}), as well as a set of neutral categories (e.g. Punctuation, Symbols, Numbers, etc.). After filtering out non-official scripts, we run a post-hoc clean-up filter that removes trailing whitespace, empty bracket groups, etc. Under this setup, foreign-script named entities and references are removed, and ones in the official script(s) are retained.

\section{Language-specific Heuristic Thresholding}\label{app:data}

Filtering Wikipedia using heuristics requires an arbitrary threshold to serve as a cut-off between `good' and `bad' quality. Existing thresholds (e.g. Gopher \cite{rae2021scaling}) are normally tuned for English, and not directly applicable to other languages. Though we attempted to filter Wikipedia articles on thresholds defined by language-specific data distributions and conducted similar downstream experiments as detailed in \S \ref{sec:downstream}, we found little gains over simple script filtering and deduplication. We detail the procedure for creating language-specific thresholds for heuristic metrics below:

For a given Wikipedia metric distribution (e.g., \texttt{rps\_doc\_word\_count}), we determine a representative sample size based on the full dataset. Testing across various metrics and Wikipedias, we select a sample size of $n_{\mathrm{sample}} = 0.05 * n_{\mathrm{docs}}$, where $n_{\mathrm{docs}}$ is the total number of documents in the dataset. We then sort the full distribution in ascending order and take the first $n_{\mathrm{sample}}$ values as the \textit{value distribution}, $D_{\mathrm{value}}$. This distribution reflects the lowest values in the distribution ---essentially, the values we want to assess as potential outliers or benchmarks. Separately, we randomly sample $n_{\mathrm{sample}}$ values from the full distribution to create the \textit{sampled distribution}, $D_{\mathrm{sample}}$. This random sample provides a baseline density that represents the overall data distribution rather than just extreme values. We estimate the kernel density (KDE) for both $D_{\mathrm{value}}$ and $D_{\mathrm{sample}}$, applying these KDEs over $n_{\mathrm{sample}}$ evenly spaced values in the range:
$$
\left[ \min(D_{\mathrm{value}}), \, \max(D_{\mathrm{sample}}) \right]
$$
The threshold is then identified as the point within this range where the difference between the two density estimates is maximized. 

\section{Training Hyperparameters and Finetuning Results}\label{app:hypers}
Hyperparameters for the pretraining and fine-tuning of monolingual models are in Table \ref{tab:tinybert-hypers}. We use the SentencePiece package \cite{kudo-richardson-2018-sentencepiece} to train monolingual unigram tokenisers for each language. We train two tokenisers per language: one on the \expRaw with no filtering, which is used to train the \expRaw and \expRandom models; the second one on the data remaining after script filtering and deduplication, which we use to train the \expPrimary models. We use an approximation with Zipf's law \cite{zipf_human_1949} on each Wikipedia to estimate the vocabulary size for each model, mentioned in Table \ref{tab:vocab}. Hyperparameters for XLM-R language adaptation are in Table \ref{tab:xlmr-hypers}. For both cases, we checkpoint and use the best model on a held-out validation set to fine-tune after pretraining.
For multilingual pretraining, we train 2 WordPiece tokenizers on \expRaw and \expPrimary with a vocabulary size of 110k. Languages are listed in Table \ref{tab:top100} and hyperparameters in Table \ref{tab:mbert-hypers}.

\begin{table}[h]
    \centering
    \small

\begin{tabular}{lrr}\toprule
\multicolumn{2}{c}{\textbf{Tiny DeBERTa Hyperparameters}} \\\cmidrule{1-2}
\multicolumn{2}{c}{\textbf{Pretraining}} \\\midrule
& \\
Model Type &deberta \\
Hidden Size &312 \\
Intermediate Size &1200 \\
Attention Heads &12 \\
Hidden Layers &4 \\
Hidden Act &gelu \\
Max Length &128 \\
Task &MLM \\
Mask Probability &0.4 \\
Padding Strategy &longest \\
Grad. Accumulation Steps &4 \\
Batch size &8 to 32 \\
Learning Rate &1.00e-3 \\
Epochs &100 \\\midrule
\multicolumn{2}{c}{\textbf{Finetuning}} \\\midrule
Max Length &128 \\
Epochs &20 \\
Batch Size &8 \\
Learning Rate &5.00e-5 \\
Padding Strategy &max\_length \\
\bottomrule
\end{tabular}
    \caption{\footnotesize Hyperparameters for monolingual pretraining and fine-tuning. While we used 100 epochs as a default, for Wikipedias that are too large (\texttt{nl, ro, id, hu, uk, vi, ceb, war, tt, azb, min}) we pretrain only for 2 epochs. Similarly, while we used a batch size of 8 for all models, for the larger Wikipedias we use a batch size of 32. All models are trained on 1 GPU.}
    \label{tab:tinybert-hypers}
\end{table}

\begin{table}
    \centering
    \small
    \begin{tabular}{lrr}\toprule
\multicolumn{2}{c}{\textbf{XLMR Hyperparameters}} \\\cmidrule{1-2}
\multicolumn{2}{c}{\textbf{Language Adaptation}} \\\midrule
Model Type &xlm-roberta \\
Task &MLM \\
Max Length &512 \\
Mask Probability &0.15 \\
Epochs &3 \\
Batch Size &8 \\
Learning Rate &5.00e-5 \\
Padding strategy &Longest \\
Grad. Accumulation Steps &1 \\\midrule
\multicolumn{2}{c}{\textbf{Finetuning - SIB200}} \\\midrule
Max Length &512 \\
Epochs &20 \\
Batch Size &32 \\
Learning Rate &5.00e-5 \\
Padding Strategy &max\_length \\\midrule
\multicolumn{2}{c}{\textbf{Finetuning - Belebele}} \\\midrule
Max Length &128 \\
Epochs &3 \\
Batch Size &8 \\
Learning Rate &5.00e-5 \\
Padding Strategy &max\_length \\
\bottomrule
\end{tabular}
    \caption{\footnotesize Hyperparameters for XLM-R language adaptation (based on \citet{alabi-etal-2022-adapting}) and fine-tuning. }
    \label{tab:xlmr-hypers}
\end{table}

\begin{table}
    \centering
    \small
    \begin{tabular}{lrr}\toprule
\multicolumn{2}{c}{\textbf{Multilingual Pretraining}} \\\cmidrule{1-2}
\multicolumn{2}{c}{\textbf{Pretraining}} \\\midrule
Sampling Temperature &1.43 \\
Model Type &bert \\
Task &MLM \\
Max Length &128 \\
Mask Probability &0.12 \\
Steps &200,000 \\
Batch Size (per device) &256 \\
Learning Rate &1.00e-3 \\
Padding strategy &Longest \\
Grad. Accumulation Steps &4 \\\midrule
\multicolumn{2}{c}{\textbf{Finetuning - Belebele}} \\\midrule
Max Length &128 \\
Epochs &3 \\
Batch Size &8 \\
Learning Rate &5.00e-5 \\
Padding Strategy &max\_length \\\midrule
\multicolumn{2}{c}{\textbf{Finetuning - MNLI}} \\\midrule
Max Length &128 \\
Epochs &3 \\
Batch Size &32 \\
Learning Rate &2.00e-5 \\
Padding Strategy &max\_length \\
\bottomrule
\end{tabular}
    \caption{\footnotesize Hyperparameters for multilingual pretraining and fine-tuning, based on mBERT and \citet{izsak-etal-2021-train}. Both models are trained on 4 H100 GPUs.}
    \label{tab:mbert-hypers}
\end{table}

\begin{table}
    \centering
    \small
    \begin{tabular}{lrrrr}\toprule
\multicolumn{4}{c}{\textbf{Estimated Vocabulary Size}} \\\cmidrule{1-4}
\textbf{Lang} &\textbf{Vocab} &\textbf{Wiki-ID} &\textbf{Vocab} \\\midrule
ace &3505 &mr &14708 \\
am &5234 &nl &60011 \\
ary &4502 &nso &2801 \\
awa &2190 &ny &2472 \\
azb &13996 &om &3436 \\
ban &6624 &pag &2170 \\
bm &1485 &pcm &2595 \\
ceb &73124 &rn &1446 \\
ckb &11451 &ro &36329 \\
ee &1770 &rw &5725 \\
fon &1426 &shn &5784 \\
ha &13126 &sn &5227 \\
ht &10836 &so &6117 \\
hu &46205 &sw &12529 \\
id &41725 &te &23175 \\
ig &11609 &ti &1150 \\
kk &22496 &tn &3783 \\
ks &2248 &ts &1817 \\
lg &4528 &tt &26047 \\
lij &5462 &tw &4745 \\
ln &2509 &uk &62626 \\
mai &4793 &vi &43806 \\
mi &3451 &war &25803 \\
min &14783 &wo &3384 \\
mni &3351 &yo &6128 \\
& &zu &4439 \\
\bottomrule
\end{tabular}
    \caption{\footnotesize Estimated vocabulary size used to train monolingual tokenisers for each Wikipedia in our experimental pipeline.}
    \label{tab:vocab}
\end{table}

\begin{table}
    \centering
    \tiny
    \begin{tabular}{lrlrr}\toprule
\multicolumn{4}{c}{\textbf{Languages Included in Multilingual Pretraining}} \\\cmidrule{1-4}
\textbf{Language} &\textbf{Wiki-ID} &\textbf{Language} &\textbf{Wiki-ID} \\\midrule
Afrikaans &af &Korean &ko \\
Albanian &sq &Latin &la \\
Arabic &ar &Latvian &lv \\
Aragonese &an &Lithuanian & lt \\
Armenian &hy &Lombard &lmo \\
Asturian &ast &Low German &nds-nl \\
Azerbaijani &az &Luxembourgish &lb \\
Bashkir &ba &Macedonian &mk \\
Basque &eu &Malagasy &mg \\
Bavarian &bar &Malay &ms \\
Belarusian &be &Malayalam &ml \\
Bangla &bn &Marathi &mr \\
Bishnupriya &bpy &Minangkabau &min \\
Bosnian &bs &Nepali &ne \\
Breton &br &Newari &new \\
Bulgarian &bg &Norwegian Nynorsk &nn \\
Burmese &my &Occitan &oc \\
Catalan &ca &Persian &fa \\
Chechen &ce &Piedmontese &pms \\
Chinese &zh &Polish &pl \\
Cantonese &zh-yue &Portuguese &pt \\
Cebuano &ceb &Punjabi &pa \\
Chuvash &cv &Romanian &ro \\
Croatian &hr &Rusyn &rue \\
Czech &cs &Scots &sco \\
Danish &da &Serbo-Croatian &sh \\
Dutch &nl &Sicilian &scn \\
English &en &Slovak &sk \\
Estonian &et &Slovenian &sl \\
Finnish &fi &South Azerbaijani &azb \\
French &fr &Spanish &es \\
Galician &gl &Sundanese &su \\
Georgian &ka &Swahili &sw \\
German &de &Swedish &sv \\
Greek &el &Filipino &tl \\
Gujarati &gu &Tajik &tg \\
Haitian Creole &ht &Tamil &ta \\
Hebrew &he &Tatar &tt \\
Hindi &hi &Turkish &tr \\
Hungarian &hu &Ukrainian &uk \\
Icelandic &is &Uzbek &uz \\
Ido &io &Vietnamese &vi \\
Indonesian &id &Volapük &vo \\
Irish &ga &Waray &war \\
Italian &it &Welsh &cy \\
Japanese &ja &Western Frisian &fy \\
Javanese &jv &Western Panjabi &pnb \\
Kannada &kn &Yoruba &yo \\
Kazakh &kk &Thai &th \\
Kyrgyz &ky &Mongolian &mn \\
\bottomrule
\end{tabular}
    \caption{\footnotesize Wikipedias included in the pretraining of the multilingual models.}
    \label{tab:top100}
\end{table}

\begin{table}
\centering
    \scriptsize

\begin{tabular}{lrrrr}\toprule
\multicolumn{4}{c}{\textbf{SIB200}} \\\cmidrule{1-4}
\textbf{wiki} &\textbf{raw\_wiki} &\textbf{+filters} &\textbf{random} \\\cmidrule{1-4}
\multicolumn{4}{c}{\textsc{Tier 1}} \\\midrule
ti &0.38$_{\pm0.05}$ &0.38$_{\pm0.01}$ &0.32$_{\pm0.01}$ \\
ha &0.77$_{\pm0.02}$ &\textbf{0.78$_{\pm0.02}$} &0.62$_{\pm0.03}$ \\
ts &\textbf{0.61$_{\pm0.03}$} &0.60$_{\pm0.01}$ &0.53$_{\pm0.02}$ \\
ig &0.76$_{\pm0.01}$ &\textbf{0.77$_{\pm0.02}$} &0.64$_{\pm0.02}$ \\
sn &\textbf{0.66$_{\pm0.03}$} &0.62$_{\pm0.03}$ &0.43$_{\pm0.05}$ \\
so &0.68$_{\pm0.01}$ &\textbf{0.73$_{\pm0.02}$} &0.56$_{\pm0.06}$ \\
rw &0.77$_{\pm0.01}$ &\textbf{0.77$_{\pm0.01}$} &0.65$_{\pm0.03}$ \\
tw &\textbf{0.65$_{\pm0.10}$} &0.65$_{\pm0.02}$ &0.63$_{\pm0.03}$ \\
ln &\textbf{0.59$_{\pm0.02}$} &0.57$_{\pm0.02}$ &0.52$_{\pm0.04}$ \\
wo &0.50$_{\pm0.02}$ &\textbf{0.52$_{\pm0.02}$} &0.45$_{\pm0.04}$ \\
om &0.59$_{\pm0.01}$ &\textbf{0.62$_{\pm0.01}$} &0.34$_{\pm0.05}$ \\
ny &\textbf{0.64$_{\pm0.02}$} &0.60$_{\pm0.01}$ &0.60$_{\pm0.01}$ \\
ee &0.59$_{\pm0.03}$ &\textbf{0.61$_{\pm0.03}$} &0.57$_{\pm0.01}$ \\
bm &0.45$_{\pm0.02}$ &\textbf{0.48$_{\pm0.02}$} &0.47$_{\pm0.03}$ \\
fon &0.48$_{\pm0.03}$ &0.45$_{\pm0.04}$ &0.44$_{\pm0.01}$ \\\midrule
\textbf{Avg} &0.61 &\textbf{0.61} & 0.51\\\midrule
\multicolumn{4}{c}{\textsc{Tier 2}} \\\midrule
ro &\textbf{0.82$_{\pm0.03}$} &0.80$_{\pm0.02}$ &0.66$_{\pm0.01}$ \\
id &0.83$_{\pm0.01}$ &\textbf{0.85$_{\pm0.01}$} &0.75$_{\pm0.04}$ \\
hu &0.82$_{\pm0.01}$ &\textbf{0.86$_{\pm0.01}$} &0.64$_{\pm0.05}$ \\
nl &\textbf{0.82$_{\pm0.03}$} &0.80$_{\pm0.03}$ &0.66$_{\pm0.03}$ \\
uk &\textbf{0.85$_{\pm0.01}$} &0.81$_{\pm0.02}$ &0.43$_{\pm0.08}$ \\
awa &0.54$_{\pm0.03}$ &0.52$_{\pm0.03}$ &0.47$_{\pm0.02}$ \\
ks &0.42$_{\pm0.02}$ &\textbf{0.45$_{\pm0.02}$} &0.41$_{\pm0.05}$ \\
ckb &0.80$_{\pm0.01}$ &\textbf{0.82$_{\pm0.01}$} &0.54$_{\pm0.04}$ \\
rn &0.51$_{\pm0.03}$ &0.50$_{\pm0.02}$  &0.45$_{\pm0.03}$ \\
sw &0.79$_{\pm0.02}$ &0.81$_{\pm0.02}$  &0.58$_{\pm0.05}$ \\
am &\textbf{0.66$_{\pm0.01}$} &0.65$_{\pm0.02}$ &0.41$_{\pm0.02}$ \\
zu &\textbf{0.73$_{\pm0.01}$} &0.70$_{\pm0.02}$ &0.61$_{\pm0.02}$ \\
tn &0.64$_{\pm0.02}$ &0.61$_{\pm0.01}$ &0.37$_{\pm0.05}$ \\
ary &0.59$_{\pm0.00}$ &0.60$_{\pm0.01}$ &0.46$_{\pm0.02}$ \\
lg &0.55$_{\pm0.02}$ &0.55$_{\pm0.02}$ &0.51$_{\pm0.02}$ \\\midrule
\textbf{Avg} &0.69 &0.69 & 0.53 \\\midrule
\multicolumn{4}{c}{\textsc{Tier 3}} \\\midrule
lij &\textbf{0.61$_{\pm0.01}$} &0.59$_{\pm0.02}$ &0.45$_{\pm0.02}$ \\
ht &\textbf{0.70$_{\pm0.04}$} &0.69$_{\pm0.01}$ &0.36$_{\pm0.03}$ \\
te &0.80$_{\pm0.02}$ &0.79$_{\pm0.02}$ &0.52$_{\pm0.03}$ \\
mr &0.74$_{\pm0.04}$ &\textbf{0.78$_{\pm0.02}$} &0.50$_{\pm0.04}$ \\
kk &0.81$_{\pm0.02}$ &\textbf{0.83$_{\pm0.02}$} &0.74$_{\pm0.02}$ \\
yo &\textbf{0.70$_{\pm0.02}$} &0.69$_{\pm0.02}$ &0.52$_{\pm0.07}$ \\
ace &\textbf{0.27$_{\pm0.02}$} &0.15$_{\pm0.03}$ &0.23$_{\pm0.01}$ \\
shn &0.74$_{\pm0.02}$ &0.72$_{\pm0.02}$ &0.62$_{\pm0.04}$ \\
mai &0.67$_{\pm0.01}$ &0.68$_{\pm0.02}$ &0.55$_{\pm0.02}$ \\
ban &0.65$_{\pm0.01}$ &0.66$_{\pm0.02}$ &0.49$_{\pm0.03}$ \\
vi &0.82$_{\pm0.01}$ &\textbf{0.83$_{\pm0.00}$} &0.74$_{\pm0.02}$ \\
nso &0.50$_{\pm0.02}$ &\textbf{0.53$_{\pm0.01}$} &0.46$_{\pm0.04}$ \\\midrule
\textbf{Avg} &\textbf{0.67} &0.66 &0.52 \\\midrule
\multicolumn{4}{c}{\textsc{Tier 4}} \\\midrule
tt &\textbf{0.79$_{\pm0.01}$} &0.75$_{\pm0.01}$ &0.48$_{\pm0.03}$ \\
ceb &0.39$_{\pm0.03}$ &\textbf{0.51$_{\pm0.03}$} &0.28$_{\pm0.04}$ \\
war &0.48$_{\pm0.04}$ &0.49$_{\pm0.04}$ &0.46$_{\pm0.02}$ \\
pag &0.45$_{\pm0.03}$ &\textbf{0.48$_{\pm0.01}$} &0.35$_{\pm0.02}$ \\
mi &\textbf{0.55$_{\pm0.02}$} &0.52$_{\pm0.02}$ &0.48$_{\pm0.03}$ \\
mni &\textbf{0.27$_{\pm0.04}$} &0.21$_{\pm0.01}$ &0.31$_{\pm0.00}$ \\
azb &\textbf{0.53$_{\pm0.05}$} &0.53$_{\pm0.02}$ &0.42$_{\pm0.03}$ \\
min &\textbf{0.65$_{\pm0.03}$} &0.60$_{\pm0.03}$ &0.41$_{\pm0.02}$ \\\midrule
\textbf{Avg} &\textbf{0.51} &0.51 & 0.40 \\
\bottomrule
\end{tabular}

    \caption{\footnotesize Fine-tuning results on SIB200 for tiny DeBERTa models. Results are averaged over 5 runs.}\label{tab:sib200-all}
\end{table}

\begin{table}
\centering
    \scriptsize

\begin{tabular}{lrrrr}\toprule
\multicolumn{4}{c}{\textbf{MasakhaNER 2}} \\\cmidrule{1-4}
\textbf{Lang} &\textbf{raw\_wiki} &\textbf{+filters} &\textbf{random} \\\midrule
pcm &\textbf{0.77$_{\pm0.00}$} &0.76$_{\pm0.00}$ &0.74$_{\pm0.00}$ \\
ha &\textbf{0.82$_{\pm0.00}$} &0.81$_{\pm0.00}$ &0.75$_{\pm0.01}$ \\
ig &0.85$_{\pm0.00}$ &\textbf{0.85$_{\pm0.00}$} &0.81$_{\pm0.01}$ \\
sn &\textbf{0.92$_{\pm0.00}$} &0.92$_{\pm0.00}$ &0.90$_{\pm0.00}$ \\
rw &0.76$_{\pm0.05}$ &\textbf{0.79$_{\pm0.01}$} &0.74$_{\pm0.01}$ \\
tw &0.76$_{\pm0.00}$ &\textbf{0.76$_{\pm0.01}$} &0.71$_{\pm0.00}$ \\
wo &\textbf{0.75$_{\pm0.01}$} &0.73$_{\pm0.01}$ &0.73$_{\pm0.01}$ \\
ny &0.87$_{\pm0.00}$ &\textbf{0.87$_{\pm0.00}$} &0.85$_{\pm0.00}$ \\
ee &0.81$_{\pm0.00}$ &\textbf{0.81$_{\pm0.01}$} &0.78$_{\pm0.01}$ \\
bm &\textbf{0.74$_{\pm0.01}$} &0.74$_{\pm0.01}$ &0.73$_{\pm0.00}$ \\
fon &0.74$_{\pm0.01}$ &0.74$_{\pm0.00}$ &0.71$_{\pm0.01}$ \\
sw &0.90$_{\pm0.00}$ &0.90$_{\pm0.00}$ &0.88$_{\pm0.00}$ \\\midrule
zu &\textbf{0.83$_{\pm0.01}$} &0.82$_{\pm0.01}$ &0.77$_{\pm0.01}$ \\
lg &0.84$_{\pm0.00}$ &\textbf{0.85$_{\pm0.00}$} &0.80$_{\pm0.01}$ \\
tn &\textbf{0.80$_{\pm0.01}$} &0.80$_{\pm0.00}$ &0.63$_{\pm0.01}$ \\\midrule
yo &0.84$_{\pm0.00}$ &\textbf{0.84$_{\pm0.00}$} &0.80$_{\pm0.00}$ \\\midrule
\textbf{Avg} &0.81 &0.81 &0.77 \\
\bottomrule
\end{tabular}

    \caption{\footnotesize Fine-tuning results on MasakhaNER 2. Results are averaged over 5 runs, and languages are grouped by tier.}\label{tab:ner}
\end{table}
    
\begin{table}
\centering
    \scriptsize
    % \begin{tabular}{lrrrrr}\toprule
% \multicolumn{5}{c}{\textbf{MasakhaNEWS}} \\\cmidrule{1-5}
% \textbf{wiki} &\textbf{raw\_wiki} &\textbf{+primary} &\textbf{+heuristic} &\textbf{random} \\\midrule
% pcm &0.90$_{\pm0.04}$ &\textbf{0.91$_{\pm0.00}$} &0.87$_{\pm0.04}$ &0.88$_{\pm0.02}$ \\
% ha &\textbf{0.88$_{\pm0.01}$} &0.86$_{\pm0.01}$ &0.86$_{\pm0.01}$ &0.84$_{\pm0.01}$ \\
% ig &\textbf{0.85$_{\pm0.01}$} &0.85$_{\pm0.01}$ &0.84$_{\pm0.01}$ &0.81$_{\pm0.01}$ \\
% sn &0.89$_{\pm0.01}$ &\textbf{0.90$_{\pm0.01}$} &0.89$_{\pm0.01}$ &0.87$_{\pm0.01}$ \\
% so &0.71$_{\pm0.01}$ &\textbf{0.71$_{\pm0.01}$} &0.70$_{\pm0.01}$ &0.59$_{\pm0.03}$ \\
% ln &\textbf{0.78$_{\pm0.01}$} &0.76$_{\pm0.01}$ &0.76$_{\pm0.01}$ &0.74$_{\pm0.02}$ \\
% om &0.67$_{\pm0.30}$ &\textbf{0.82$_{\pm0.01}$} &0.81$_{\pm0.30}$ &0.71$_{\pm0.04}$ \\
% ti &0.60$_{\pm0.04}$ &\textbf{0.65$_{\pm0.02}$} &0.62$_{\pm0.04}$ &0.56$_{\pm0.02}$ \\\midrule
% sw &0.81$_{\pm0.01}$ &0.82$_{\pm0.00}$ &\textbf{0.82$_{\pm0.01}$} &0.75$_{\pm0.01}$ \\
% am &0.84$_{\pm0.04}$ &\textbf{0.87$_{\pm0.01}$} &0.85$_{\pm0.04}$ &0.83$_{\pm0.01}$ \\
% lg &0.86$_{\pm0.01}$ &\textbf{0.86$_{\pm0.02}$} &0.84$_{\pm0.01}$ &0.84$_{\pm0.02}$ \\
% rn &\textbf{0.75$_{\pm0.01}$} &0.76$_{\pm0.02}$ &0.71$_{\pm0.01}$ &0.71$_{\pm0.01}$ \\\midrule
% yo &0.84$_{\pm0.03}$ &0.86$_{\pm0.01}$ &\textbf{0.86$_{\pm0.03}$} &0.82$_{\pm0.01}$ \\\midrule
% \textbf{Avg} &0.8 &\textbf{0.82} &0.8 &0.76 \\
% \bottomrule
% \end{tabular}

\begin{tabular}{lrrrr}\toprule
\multicolumn{4}{c}{\textbf{MasakhaNEWS}} \\\cmidrule{1-4}
\textbf{wiki} &\textbf{raw\_wiki} &\textbf{+filters} &\textbf{random} \\\midrule
pcm &0.90$_{\pm0.04}$ &\textbf{0.91$_{\pm0.00}$} &0.88$_{\pm0.02}$ \\
ha &\textbf{0.88$_{\pm0.01}$} &0.86$_{\pm0.01}$ &0.84$_{\pm0.01}$ \\
ig &\textbf{0.85$_{\pm0.01}$} &0.85$_{\pm0.01}$ &0.81$_{\pm0.01}$ \\
sn &0.89$_{\pm0.01}$ &\textbf{0.90$_{\pm0.01}$} &0.87$_{\pm0.01}$ \\
so &0.71$_{\pm0.01}$ &\textbf{0.71$_{\pm0.01}$} &0.59$_{\pm0.03}$ \\
ln &\textbf{0.78$_{\pm0.01}$} &0.76$_{\pm0.01}$ &0.74$_{\pm0.02}$ \\
om &0.67$_{\pm0.30}$ &\textbf{0.82$_{\pm0.01}$} &0.71$_{\pm0.04}$ \\
ti &0.60$_{\pm0.04}$ &\textbf{0.65$_{\pm0.02}$} &0.56$_{\pm0.02}$ \\\midrule
sw &0.81$_{\pm0.01}$ &0.82$_{\pm0.00}$ &0.75$_{\pm0.01}$ \\
am &0.84$_{\pm0.04}$ &\textbf{0.87$_{\pm0.01}$} &0.83$_{\pm0.01}$ \\
lg &0.86$_{\pm0.01}$ &\textbf{0.86$_{\pm0.02}$} &0.84$_{\pm0.02}$ \\
rn &\textbf{0.75$_{\pm0.01}$} &0.76$_{\pm0.02}$ &0.71$_{\pm0.01}$ \\\midrule
yo &0.84$_{\pm0.03}$ &0.86$_{\pm0.01}$ &0.82$_{\pm0.01}$ \\\midrule
\textbf{Avg} &0.8 &\textbf{0.82} &0.76 \\
\bottomrule
\end{tabular}
    \caption{\footnotesize Fine-tuning results on MasakhaNEWS. Results are averaged over 5 runs, and languages are grouped by tier.}\label{tab:news}
\end{table}

\begin{table}
\centering
    \scriptsize
    % \begin{tabular}{lrrrrr}\toprule
% \multicolumn{5}{c}{\textbf{AfriSenti-Twitter}} \\\cmidrule{1-5}
% \textbf{wiki} &\textbf{raw\_wiki} &\textbf{+primary} &\textbf{+heuristic} &\textbf{random} \\\midrule
% pcm &0.61$_{\pm0.01}$ &\textbf{0.61$_{\pm0.01}$} &0.61$_{\pm0.01}$ &0.60$_{\pm0.01}$ \\
% ha &0.72$_{\pm0.01}$ &0.72$_{\pm0.01}$ &\textbf{0.73$_{\pm0.01}$} &0.72$_{\pm0.01}$ \\
% ig &0.74$_{\pm0.00}$ &0.75$_{\pm0.00}$ &\textbf{0.75$_{\pm0.00}$} &0.74$_{\pm0.01}$ \\
% rw &0.60$_{\pm0.04}$ &\textbf{0.61$_{\pm0.01}$} &0.59$_{\pm0.04}$ &0.60$_{\pm0.01}$ \\
% tw &0.62$_{\pm0.01}$ &\textbf{0.64$_{\pm0.01}$} &0.64$_{\pm0.01}$ &0.63$_{\pm0.01}$ \\
% ts &0.48$_{\pm0.04}$ &0.49$_{\pm0.03}$ &\textbf{0.50$_{\pm0.04}$} &\textbf{0.50$_{\pm0.01}$} \\
% sw &0.55$_{\pm0.03}$ &\textbf{0.59$_{\pm0.02}$} &0.56$_{\pm0.03}$ &0.53$_{\pm0.04}$ \\\midrule
% am &0.40$_{\pm0.07}$ &0.41$_{\pm0.04}$ &\textbf{0.46$_{\pm0.07}$} &0.45$_{\pm0.10}$ \\
% ary &\textbf{0.46$_{\pm0.01}$} &0.40$_{\pm0.01}$ &0.40$_{\pm0.01}$ &0.44$_{\pm0.01}$ \\\midrule
% yo &0.68$_{\pm0.01}$ &\textbf{0.68$_{\pm0.01}$} &0.68$_{\pm0.01}$ &0.64$_{\pm0.01}$ \\\midrule
% \textbf{Avg} &0.59 &\textbf{0.59} &0.59 &0.59 \\
% \bottomrule
% \end{tabular}

\begin{tabular}{lrrrr}\toprule
\multicolumn{4}{c}{\textbf{AfriSenti-Twitter}} \\\cmidrule{1-4}
\textbf{wiki} &\textbf{raw\_wiki} &\textbf{+filters} &\textbf{random} \\\midrule
pcm &0.61$_{\pm0.01}$ &\textbf{0.61$_{\pm0.01}$} &0.60$_{\pm0.01}$ \\
ha &0.72$_{\pm0.01}$ &0.72$_{\pm0.01}$ &0.72$_{\pm0.01}$ \\
ig &0.74$_{\pm0.00}$ &0.75$_{\pm0.00}$ &0.74$_{\pm0.01}$ \\
rw &0.60$_{\pm0.04}$ &\textbf{0.61$_{\pm0.01}$} &0.60$_{\pm0.01}$ \\
tw &0.62$_{\pm0.01}$ &\textbf{0.64$_{\pm0.01}$} &0.63$_{\pm0.01}$ \\
ts &0.48$_{\pm0.04}$ &0.49$_{\pm0.03}$ &\textbf{0.50$_{\pm0.01}$} \\
sw &0.55$_{\pm0.03}$ &\textbf{0.59$_{\pm0.02}$} &0.53$_{\pm0.04}$ \\\midrule
am &0.40$_{\pm0.07}$ &0.41$_{\pm0.04}$ &0.45$_{\pm0.10}$ \\
ary &\textbf{0.46$_{\pm0.01}$} &0.40$_{\pm0.01}$ &0.44$_{\pm0.01}$ \\\midrule
yo &0.68$_{\pm0.01}$ &\textbf{0.68$_{\pm0.01}$} &0.64$_{\pm0.01}$ \\\midrule
\textbf{Avg} &0.59 &\textbf{0.59} &0.59 \\
\bottomrule
\end{tabular}
    \caption{\footnotesize Fine-tuning results on AfriSenti-Twitter. Results are averaged over 5 runs, and languages are grouped by tier.}\label{tab:sent}
\end{table}

% \subsection{Language Adaptation}
\begin{table}
\centering
\scriptsize

\begin{tabular}{lrrrr}\toprule
\multicolumn{4}{c}{\textbf{XLMR Continued Pretraining (SIB200)}} \\\cmidrule{1-4}
\textbf{wiki} &\textbf{baseline} &\textbf{raw\_wiki} &\textbf{+filters} & \\\cmidrule{1-4}
ti &0.59$_{\pm0.04}$ &0.53$_{\pm0.21}$ &0.56$_{\pm0.02}$ \\
fon &0.43$_{\pm0.20}$ &\textbf{0.60$_{\pm0.04}$} &0.56$_{\pm0.02}$ \\
ig &0.52$_{\pm0.20}$ &\textbf{0.79}$_{\pm0.02}$ &\textbf{0.79$_{\pm0.03}$} \\
ee &0.64$_{\pm0.04}$ &\textbf{0.66$_{\pm0.03}$} &0.64$_{\pm0.03}$ \\
ny &0.64$_{\pm0.01}$ &\textbf{0.68}$_{\pm0.03}$ &0.65$_{\pm0.02}$ \\
wo &0.60$_{\pm0.03}$ &\textbf{0.65}$_{\pm0.01}$ &0.64$_{\pm0.01}$ \\
ln &0.66$_{\pm0.01}$ &\textbf{0.70$_{\pm0.01}$} &0.69$_{\pm0.01}$ \\
tw &0.66$_{\pm0.02}$ &\textbf{0.72$_{\pm0.02}$} &0.70$_{\pm0.03}$ \\
rw &0.56$_{\pm0.03}$ &\textbf{0.69$_{\pm0.02}$} &0.67$_{\pm0.02}$ \\
sn &0.54$_{\pm0.03}$ &0.59$_{\pm0.02}$ &\textbf{0.60}$_{\pm0.03}$ \\
bm &0.43$_{\pm0.22}$ &\textbf{0.60$_{\pm0.05}$} &0.55$_{\pm0.03}$ \\
ts &0.58$_{\pm0.05}$ &\textbf{0.63$_{\pm0.04}$} &0.60$_{\pm0.05}$ \\\cmidrule{1-4}
\textbf{Avg} &0.57 &0.65 &0.64 \\\cmidrule{1-4}
tn &0.56$_{\pm0.02}$ &\textbf{0.62$_{\pm0.03}$} &0.60$_{\pm0.04}$ \\
ks &0.62$_{\pm0.22}$ &\textbf{0.68$_{\pm0.02}$} &\textbf{0.68}$_{\pm0.02}$ \\
rn &0.54$_{\pm0.04}$ &0.55$_{\pm0.06}$ &\textbf{0.58$_{\pm0.02}$} \\
lg &0.40$_{\pm0.17}$ &\textbf{0.58$_{\pm0.03}$} &\textbf{0.58}$_{\pm0.03}$ \\
zu &0.58$_{\pm0.01}$ &0.64$_{\pm0.03}$ &\textbf{0.68$_{\pm0.02}$} \\
awa &0.80$_{\pm0.04}$ &0.81$_{\pm0.02}$ &\textbf{0.83$_{\pm0.03}$} \\\cmidrule{1-4}
\textbf{Avg} &0.59 &0.65 &0.66 \\\cmidrule{1-4}
ht &0.63$_{\pm0.03}$ &\textbf{0.73$_{\pm0.03}$} &\textbf{0.73}$_{\pm0.02}$ \\
nso &0.55$_{\pm0.02}$ &0.59$_{\pm0.04}$ &\textbf{0.63$_{\pm0.02}$} \\
lij &0.77$_{\pm0.02}$ &\textbf{0.81$_{\pm0.03}$} &0.78$_{\pm0.03}$ \\
ace &0.41$_{\pm0.09}$ &0.45$_{\pm0.13}$ &\textbf{0.53$_{\pm0.06}$} \\
shn &0.33$_{\pm0.16}$ &0.61$_{\pm0.03}$ &0.60$_{\pm0.04}$ \\
mai &0.82$_{\pm0.02}$ &0.83$_{\pm0.02}$ &0.82$_{\pm0.02}$ \\
ban &0.79$_{\pm0.01}$ &0.82$_{\pm0.02}$ &\textbf{0.83$_{\pm0.01}$} \\
yo &0.56$_{\pm0.02}$ &0.63$_{\pm0.04}$ &\textbf{0.66$_{\pm0.02}$} \\\cmidrule{1-4}
\textbf{Avg} &0.61 &0.68 &0.70 \\\cmidrule{1-4}
mi &0.59$_{\pm0.12}$ &0.62$_{\pm0.02}$ &\textbf{0.63$_{\pm0.03}$} \\
mni &\textbf{0.54$_{\pm0.05}$} &0.38$_{\pm0.11}$ &0.45$_{\pm0.03}$ \\
pag &0.77$_{\pm0.02}$ &\textbf{0.79$_{\pm0.02}$} &0.77$_{\pm0.05}$ \\\cmidrule{1-4}
\textbf{Avg} &0.59 &0.66 &0.66 \\
\bottomrule
\end{tabular}
    \caption{\footnotesize XLM-R base and XLM-R with continued pretraining on unseen languages with \expRaw and \expPrimary fine-tuned on \textsc{SIB200}. Results are grouped by tier.}\label{tab:xlmr}
\end{table}

\begin{table}
\centering
\scriptsize
    \begin{tabular}{lrrrr}\toprule
\multicolumn{4}{c}{\textbf{XLMR Continued Pretraining (Belebele)}} \\\cmidrule{1-4}
\textbf{wiki} &\textbf{baseline} &\textbf{raw\_wiki} &\textbf{+filters} \\\midrule
wo &0.27 &0.26 &\textbf{0.29} \\
ts &0.24 &0.25 &\textbf{0.33} \\
ny &0.26 &0.25 &\textbf{0.27} \\
rw &0.24 &0.23 &\textbf{0.30} \\
ig &0.26 &0.25 &\textbf{0.30} \\
ln &0.25 &0.24 &\textbf{0.30} \\
sn &0.25 &0.23 &\textbf{0.33} \\
bm &0.24 &\textbf{0.26} &0.24 \\
ti &0.24 &0.27 &\textbf{0.29} \\\midrule
lg &0.24 &0.24 &\textbf{0.28} \\
tn &0.25 &0.24 &\textbf{0.33} \\
zu &0.25 &0.23 &\textbf{0.32} \\\midrule
shn &0.25 &0.26 &\textbf{0.27} \\
nso &0.25 &0.26 &\textbf{0.31} \\
ht &0.24 &0.24 &\textbf{0.33} \\
yo &0.26 &0.25 &\textbf{0.27} \\\midrule
mi &0.26 &0.27 &\textbf{0.29} \\\midrule
\textbf{Avg} &0.25 &0.25 &\textbf{0.30} \\
\bottomrule
\end{tabular}
    \caption{\footnotesize XLM-R base and XLM-R with continued pretraining on unseen languages with \expRaw and \expPrimary fine-tuned on \textsc{MC} and tested on \textsc{Belebele}. Languages are grouped by tier.}\label{tab:belebele-xlmr}
\end{table}

% \subsection{Multilingual Pretraining}

\begin{table}
\centering
\scriptsize
    \begin{tabular}{lrrr}\toprule
\multicolumn{3}{c}{\textbf{Multilingual Pretraining (XNLI)}} \\\cmidrule{1-3}
\textbf{wiki} &\textbf{raw\_wiki} &\textbf{+filters} \\\midrule
\textbf{el} &0.39 &0.39 \\
\textbf{bg} &0.42 &0.40 \\
\textbf{de} &0.46 &0.43 \\
\textbf{tr} &0.43 &0.41 \\
\textbf{th} &0.38 &0.37 \\
\textbf{fr} &0.48 &0.46 \\
\textbf{ru} &0.40 &0.38 \\
\textbf{es} &0.50 &0.49 \\
\textbf{zh} &0.43 &0.43 \\
\textbf{en} &0.72 &0.71 \\\midrule
\textbf{ar} &0.39 &0.38 \\
\textbf{sw} &0.41 &0.39 \\
\textbf{ur} &0.38 &0.39 \\
\textbf{hi} &0.38 &0.37 \\\midrule
\textbf{vi} &0.41 &0.41 \\\midrule
\textbf{Avg} &0.44 &0.43 \\
\bottomrule
\end{tabular}
    \caption{\footnotesize Multilingual models pretrained on top 100 Wikipedias with \expRaw and \expPrimary fine-tuned on \textsc{MNLI} and tested on \textsc{XNLI}. Languages are grouped by tier.}\label{tab:xnli}
\end{table}

\begin{table}
\centering
\scriptsize
    \begin{tabular}{lrrr}\toprule
\multicolumn{3}{c}{\textbf{Multilingual Pretraining (Belebele)}} \\\midrule
\textbf{wiki} &\textbf{raw\_wiki} &\textbf{+filters} \\\midrule
\textbf{pa} &0.27 &0.26 \\
\textbf{ta} &0.24 &0.23 \\
\textbf{af} &0.24 &0.23 \\
\textbf{th} &0.22 &0.25 \\
\textbf{fr} &0.24 &0.26 \\
\textbf{es} &0.22 &0.26 \\
\textbf{cs} &0.26 &0.24 \\
\textbf{pl} &0.26 &0.24 \\
\textbf{de} &0.27 &0.26 \\
\textbf{en} &0.25 &0.26 \\
\textbf{fi} &0.26 &0.25 \\
\textbf{pt} &0.25 &0.27 \\
\textbf{hr} &0.25 &0.27 \\
\textbf{el} &0.24 &0.23 \\
\textbf{da} &0.26 &0.25 \\
\textbf{it} &0.25 &0.26 \\
\textbf{bn} &0.25 &0.27 \\
\textbf{ml} &0.27 &0.24 \\
\textbf{is} &0.25 &0.27 \\
\textbf{bg} &0.27 &0.24 \\
\textbf{kn} &0.26 &0.25 \\
\textbf{et} &0.27 &0.26 \\
\textbf{ca} &0.28 &0.26 \\
\textbf{ko} &0.27 &0.20 \\
\textbf{tr} &0.24 &0.26 \\
\textbf{he} &0.27 &0.24 \\
\textbf{zh} &0.25 &0.24 \\
\textbf{sl} &0.25 &0.25 \\
\textbf{ja} &0.27 &0.24 \\
\textbf{lt} &0.26 &0.23 \\\midrule
\textbf{sk} &0.23 &0.24 \\
\textbf{ky} &0.24 &0.26 \\
\textbf{ka} &0.25 &0.25 \\
\textbf{mk} &0.27 &0.25 \\
\textbf{uk} &0.24 &0.26 \\
\textbf{hu} &0.25 &0.27 \\
\textbf{ro} &0.25 &0.27 \\
\textbf{id} &0.26 &0.27 \\
\textbf{hi} &0.26 &0.24 \\
\textbf{jv} &0.23 &0.24 \\
\textbf{hy} &0.26 &0.24 \\
\textbf{my} &0.23 &0.24 \\
\textbf{nl} &0.26 &0.23 \\\midrule
\textbf{sv} &0.22 &0.26 \\
\textbf{yo} &0.23 &0.26 \\
\textbf{vi} &0.24 &0.26 \\
\textbf{tg} &0.24 &0.24 \\
\textbf{su} &0.24 &0.28 \\
\textbf{gu} &0.23 &0.24 \\
\textbf{eu} &0.26 &0.25 \\
\textbf{ht} &0.26 &0.26 \\
\textbf{kk} &0.27 &0.26 \\
\textbf{mr} &0.28 &0.24 \\\midrule
\textbf{ceb} &0.25 &0.26 \\
\textbf{war} &0.24 &0.23 \\\midrule
\textbf{Avg} &0.25 &0.25 \\
\bottomrule
\end{tabular}
    \caption{\footnotesize Multilingual models pretrained on top 100 Wikipedias with \expRaw and \expPrimary fine-tuned on \textsc{MC} and tested on \textsc{Belebele}. Languages are grouped by tier.}\label{tab:belebele-mbert}
\end{table}

\onecolumn
\section{Quality Tiers}\label{app:tiers}

\begin{table}[h]
    \centering
    \tiny
    \begin{tabular}{lrrrrrrrrrrrr}\toprule
\textbf{Wiki-Id} &\textbf{Tier} &\textbf{Wiki-Id} &\textbf{Tier} &\textbf{Wiki-Id} &\textbf{Tier} &\textbf{Wiki-Id} &\textbf{Tier} &\textbf{Wiki-Id} &\textbf{Tier} &\textbf{Wiki-Id} &\textbf{Tier} \\\midrule
ab &\textsc{Tier 3} &crh &\textsc{Tier 3} &ha &\textsc{Tier 1} &lez &\textsc{Tier 1} &pa &\textsc{Tier 1} &su &\textsc{Tier 3} \\
ace &\textsc{Tier 3} &cs &\textsc{Tier 1} &hak &\textsc{Tier 3} &lfn &\textsc{Tier 1} &pag &\textsc{Tier 4} &sv &\textsc{Tier 3} \\
ady &\textsc{Tier 1} &csb &\textsc{Tier 1} &haw &\textsc{Tier 3} &lg &\textsc{Tier 2} &pam &\textsc{Tier 2} &sw &\textsc{Tier 2} \\
af &\textsc{Tier 1} &cu &\textsc{Tier 1} &he &\textsc{Tier 1} &li &\textsc{Tier 1} &pap &\textsc{Tier 1} &szl &\textsc{Tier 3} \\
als &\textsc{Tier 1} &cv &\textsc{Tier 4} &hi &\textsc{Tier 2} &lij &\textsc{Tier 3} &pcd &\textsc{Tier 1} &ta &\textsc{Tier 1} \\
alt &\textsc{Tier 2} &cy &\textsc{Tier 2} &hif &\textsc{Tier 2} &lld &\textsc{Tier 4} &pcm &\textsc{Tier 1} &tay &\textsc{Tier 2} \\
am &\textsc{Tier 2} &da &\textsc{Tier 1} &hr &\textsc{Tier 1} &lmo &\textsc{Tier 3} &pdc &\textsc{Tier 1} &tcy &\textsc{Tier 1} \\
ami &\textsc{Tier 2} &dag &\textsc{Tier 3} &hsb &\textsc{Tier 2} &ln &\textsc{Tier 1} &pfl &\textsc{Tier 3} &te &\textsc{Tier 3} \\
an &\textsc{Tier 1} &de &\textsc{Tier 1} &ht &\textsc{Tier 3} &lo &\textsc{Tier 2} &pi &\textsc{Tier 4} &tet &\textsc{Tier 1} \\
ang &\textsc{Tier 1} &din &\textsc{Tier 1} &hu &\textsc{Tier 2} &lt &\textsc{Tier 1} &pih &\textsc{Tier 1} &tg &\textsc{Tier 3} \\
ar &\textsc{Tier 2} &diq &\textsc{Tier 3} &hy &\textsc{Tier 2} &ltg &\textsc{Tier 1} &pl &\textsc{Tier 1} &th &\textsc{Tier 1} \\
arc &\textsc{Tier 1} &dsb &\textsc{Tier 1} &hyw &\textsc{Tier 1} &lv &\textsc{Tier 1} &pms &\textsc{Tier 3} &ti &\textsc{Tier 1} \\
arz &\textsc{Tier 3} &dty &\textsc{Tier 2} &ia &\textsc{Tier 3} &mad &\textsc{Tier 1} &pnb &\textsc{Tier 2} &tk &\textsc{Tier 2} \\
as &\textsc{Tier 1} &dv &\textsc{Tier 1} &id &\textsc{Tier 2} &mai &\textsc{Tier 3} &pnt &\textsc{Tier 2} &tl &\textsc{Tier 1} \\
ast &\textsc{Tier 2} &dz &\textsc{Tier 1} &ie &\textsc{Tier 3} &map-bms &\textsc{Tier 3} &ps &\textsc{Tier 1} &tly &\textsc{Tier 3} \\
atj &\textsc{Tier 2} &ee &\textsc{Tier 1} &ig &\textsc{Tier 1} &mdf &\textsc{Tier 3} &pt &\textsc{Tier 1} &tn &\textsc{Tier 2} \\
av &\textsc{Tier 2} &el &\textsc{Tier 1} &ik &\textsc{Tier 1} &mg &\textsc{Tier 3} &pwn &\textsc{Tier 1} &to &\textsc{Tier 1} \\
awa &\textsc{Tier 2} &eml &\textsc{Tier 3} &ilo &\textsc{Tier 2} &mhr &\textsc{Tier 2} &qu &\textsc{Tier 2} &tpi &\textsc{Tier 2} \\
ay &\textsc{Tier 1} &eo &\textsc{Tier 1} &inh &\textsc{Tier 1} &mi &\textsc{Tier 4} &rm &\textsc{Tier 2} &tr &\textsc{Tier 1} \\
az &\textsc{Tier 2} &es &\textsc{Tier 1} &io &\textsc{Tier 2} &min &\textsc{Tier 4} &rmy &\textsc{Tier 1} &trv &\textsc{Tier 1} \\
azb &\textsc{Tier 4} &et &\textsc{Tier 1} &is &\textsc{Tier 1} &mk &\textsc{Tier 2} &rn &\textsc{Tier 2} &ts &\textsc{Tier 1} \\
ba &\textsc{Tier 2} &eu &\textsc{Tier 3} &it &\textsc{Tier 1} &ml &\textsc{Tier 1} &ro &\textsc{Tier 2} &tt &\textsc{Tier 4} \\
ban &\textsc{Tier 3} &ext &\textsc{Tier 1} &iu &\textsc{Tier 2} &mn &\textsc{Tier 1} &roa-rup &\textsc{Tier 1} &tum &\textsc{Tier 3} \\
bar &\textsc{Tier 1} &fa &\textsc{Tier 2} &ja &\textsc{Tier 1} &mni &\textsc{Tier 4} &roa-tara &\textsc{Tier 2} &tw &\textsc{Tier 1} \\
bat-smg &\textsc{Tier 3} &fat &\textsc{Tier 1} &jam &\textsc{Tier 1} &mnw &\textsc{Tier 1} &ru &\textsc{Tier 1} &ty &\textsc{Tier 3} \\
bcl &\textsc{Tier 2} &ff &\textsc{Tier 1} &jbo &\textsc{Tier 1} &mr &\textsc{Tier 3} &rue &\textsc{Tier 2} &tyv &\textsc{Tier 1} \\
be &\textsc{Tier 2} &fi &\textsc{Tier 1} &jv &\textsc{Tier 2} &mrj &\textsc{Tier 2} &rw &\textsc{Tier 1} &udm &\textsc{Tier 2} \\
be-x-old &\textsc{Tier 1} &fiu-vro &\textsc{Tier 1} &ka &\textsc{Tier 2} &ms &\textsc{Tier 3} &sa &\textsc{Tier 2} &ug &\textsc{Tier 2} \\
bg &\textsc{Tier 1} &fj &\textsc{Tier 1} &kaa &\textsc{Tier 1} &mt &\textsc{Tier 1} &sah &\textsc{Tier 1} &uk &\textsc{Tier 2} \\
bh &\textsc{Tier 2} &fo &\textsc{Tier 1} &kab &\textsc{Tier 1} &mwl &\textsc{Tier 1} &sat &\textsc{Tier 1} &ur &\textsc{Tier 2} \\
bi &\textsc{Tier 1} &fon &\textsc{Tier 1} &kbd &\textsc{Tier 1} &my &\textsc{Tier 2} &sc &\textsc{Tier 1} &uz &\textsc{Tier 2} \\
bjn &\textsc{Tier 3} &fr &\textsc{Tier 1} &kbp &\textsc{Tier 1} &myv &\textsc{Tier 2} &scn &\textsc{Tier 2} &ve &\textsc{Tier 3} \\
blk &\textsc{Tier 1} &frp &\textsc{Tier 1} &kcg &\textsc{Tier 1} &mzn &\textsc{Tier 3} &sco &\textsc{Tier 1} &vec &\textsc{Tier 3} \\
bm &\textsc{Tier 1} &frr &\textsc{Tier 1} &kg &\textsc{Tier 2} &nah &\textsc{Tier 2} &sd &\textsc{Tier 1} &vep &\textsc{Tier 1} \\
bn &\textsc{Tier 1} &fur &\textsc{Tier 1} &ki &\textsc{Tier 2} &nap &\textsc{Tier 3} &se &\textsc{Tier 2} &vi &\textsc{Tier 3} \\
bo &\textsc{Tier 2} &fy &\textsc{Tier 1} &kk &\textsc{Tier 3} &nds &\textsc{Tier 1} &sg &\textsc{Tier 1} &vls &\textsc{Tier 1} \\
bpy &\textsc{Tier 3} &ga &\textsc{Tier 1} &kl &\textsc{Tier 1} &ne &\textsc{Tier 1} &sh &\textsc{Tier 3} &vo &\textsc{Tier 2} \\
br &\textsc{Tier 1} &gag &\textsc{Tier 2} &km &\textsc{Tier 1} &new &\textsc{Tier 4} &shi &\textsc{Tier 2} &wa &\textsc{Tier 1} \\
bs &\textsc{Tier 2} &gan &\textsc{Tier 3} &kn &\textsc{Tier 1} &nia &\textsc{Tier 2} &shn &\textsc{Tier 3} &war &\textsc{Tier 4} \\
bug &\textsc{Tier 4} &gcr &\textsc{Tier 1} &ko &\textsc{Tier 1} &nl &\textsc{Tier 2} &si &\textsc{Tier 1} &wo &\textsc{Tier 1} \\
bxr &\textsc{Tier 1} &gd &\textsc{Tier 1} &koi &\textsc{Tier 3} &nn &\textsc{Tier 1} &simple &\textsc{Tier 1} &wuu &\textsc{Tier 1} \\
ca &\textsc{Tier 1} &gl &\textsc{Tier 1} &krc &\textsc{Tier 2} &no &\textsc{Tier 1} &sk &\textsc{Tier 2} &xal &\textsc{Tier 3} \\
cbk-zam &\textsc{Tier 3} &glk &\textsc{Tier 3} &ks &\textsc{Tier 2} &nov &\textsc{Tier 2} &sl &\textsc{Tier 1} &xh &\textsc{Tier 2} \\
cdo &\textsc{Tier 3} &gn &\textsc{Tier 1} &ksh &\textsc{Tier 1} &nrm &\textsc{Tier 2} &sm &\textsc{Tier 1} &xmf &\textsc{Tier 2} \\
ce &\textsc{Tier 4} &gom &\textsc{Tier 1} &ku &\textsc{Tier 3} &nso &\textsc{Tier 3} &sn &\textsc{Tier 1} &yi &\textsc{Tier 1} \\
ceb &\textsc{Tier 4} &gor &\textsc{Tier 3} &kv &\textsc{Tier 2} &nv &\textsc{Tier 3} &so &\textsc{Tier 1} &yo &\textsc{Tier 3} \\
ch &\textsc{Tier 2} &got &\textsc{Tier 2} &kw &\textsc{Tier 1} &ny &\textsc{Tier 1} &sq &\textsc{Tier 2} &za &\textsc{Tier 2} \\
chr &\textsc{Tier 3} &gu &\textsc{Tier 3} &ky &\textsc{Tier 2} &oc &\textsc{Tier 2} &sr &\textsc{Tier 3} &zea &\textsc{Tier 2} \\
chy &\textsc{Tier 1} &guc &\textsc{Tier 1} &la &\textsc{Tier 2} &olo &\textsc{Tier 1} &srn &\textsc{Tier 3} &zh &\textsc{Tier 1} \\
ckb &\textsc{Tier 2} &gur &\textsc{Tier 1} &lad &\textsc{Tier 1} &om &\textsc{Tier 1} &ss &\textsc{Tier 1} &zh-classical &\textsc{Tier 1} \\
co &\textsc{Tier 1} &guw &\textsc{Tier 1} &lb &\textsc{Tier 1} &or &\textsc{Tier 2} &st &\textsc{Tier 2} &zh-min-nan &\textsc{Tier 4} \\
cr &\textsc{Tier 3} &gv &\textsc{Tier 1} &lbe &\textsc{Tier 2} &os &\textsc{Tier 2} &stq &\textsc{Tier 1} &zh-yue &\textsc{Tier 2} \\
& & & & & & & & & &zu &\textsc{Tier 2} \\
\bottomrule
\end{tabular}
    \caption{\footnotesize Quality categorisations for all non-English Wikipedias.}
    \label{tab:tiers}
\end{table}

\newpage
\section{Heuristics per Tier}\label{app:heuristics_per_tier}

\begin{table*}[h]
    \centering
    \tiny

\begin{tabular}{@{}lc cccc @{\hspace{5pt}} lc cccc @{\hspace{5pt}} lc cccc@{}}
\toprule
\textbf{Wiki-Id} & \textbf{Len.} & \textbf{Uniq.} & \textbf{Ent.} & \textbf{W/L} & \textbf{Pct.} & \textbf{Wiki-Id} & \textbf{Len.} & \textbf{Uniq.} & \textbf{Ent.} & \textbf{W/L} & \textbf{Pct.} & \textbf{Wiki-Id} & \textbf{Len.} & \textbf{Uniq.} & \textbf{Ent.} & \textbf{W/L} & \textbf{Pct.} \\
\midrule
de & 317.00 & 176.00 & 6.88 & 11.46 & 0.30 & sah & 83.00 & 61.00 & 5.62 & 7.77 & 0.33 & ff & 43.00 & 33.00 & 4.88 & 9.83 & 0.33 \\
ru & 251.00 & 152.00 & 6.71 & 10.38 & 0.31 & gd & 92.00 & 59.00 & 5.59 & 8.50 & 0.22 & jam & 55.50 & 40.00 & 5.04 & 17.38 & 0.33 \\
es & 259.00 & 138.00 & 6.44 & 11.12 & 0.25 & sd & 107.00 & 70.00 & 5.81 & 15.67 & 0.25 & kbp & 292.00 & 146.00 & 6.56 & 26.62 & 0.38 \\
it & 168.00 & 107.00 & 6.34 & 8.00 & 0.21 & yi & 119.00 & 75.00 & 5.89 & 9.56 & 0.31 & wo & 63.00 & 44.00 & 5.25 & 6.00 & 0.25 \\
pl & 135.00 & 90.00 & 6.11 & 8.04 & 0.23 & li & 184.00 & 107.00 & 6.26 & 14.29 & 0.33 & kbd & 81.00 & 59.00 & 5.50 & 7.83 & 0.33 \\
ja & 462.00 & 203.00 & 6.93 & 11.87 & 0.27 & fo & 86.00 & 52.00 & 5.41 & 7.14 & 0.22 & nqo & 200.50 & 110.50 & 6.23 & 12.42 & 0.29 \\
pt & 149.00 & 90.00 & 6.09 & 10.43 & 0.27 & as & 358.00 & 212.00 & 7.19 & 14.71 & 0.27 & bi & 24.00 & 18.00 & 4.07 & 5.67 & 0.33 \\
ca & 248.00 & 132.00 & 6.43 & 12.86 & 0.26 & wa & 112.00 & 71.00 & 5.80 & 7.92 & 0.33 & tet & 82.00 & 58.00 & 5.56 & 7.00 & 0.33 \\
ko & 114.00 & 76.00 & 5.86 & 5.75 & 0.20 & hyw & 237.00 & 155.00 & 6.78 & 11.65 & 0.33 & roa-rup & 46.00 & 35.00 & 4.93 & 8.98 & 0.33 \\
fr & 236.00 & 129.00 & 6.45 & 8.47 & 0.24 & sn & 90.00 & 62.00 & 5.67 & 7.33 & 0.42 & jbo & 66.00 & 39.00 & 4.99 & 11.40 & 0.00 \\
no & 137.00 & 85.00 & 6.02 & 7.57 & 0.20 & co & 75.00 & 51.00 & 5.39 & 6.67 & 0.25 & fj & 27.00 & 22.00 & 4.37 & 7.00 & 0.33 \\
tr & 96.00 & 68.00 & 5.79 & 6.14 & 0.18 & so & 90.00 & 64.00 & 5.75 & 9.39 & 0.25 & guw & 174.00 & 93.00 & 6.09 & 15.54 & 0.33 \\
cs & 242.00 & 154.00 & 6.78 & 9.29 & 0.20 & vls & 161.00 & 97.00 & 6.14 & 11.16 & 0.30 & cu & 38.00 & 30.00 & 4.76 & 6.20 & 0.00 \\
eo & 130.00 & 80.00 & 5.90 & 8.69 & 0.25 & nds-nl & 166.00 & 101.00 & 6.21 & 11.29 & 0.30 & rmy & 24.00 & 21.00 & 4.33 & 7.00 & 0.40 \\
he & 393.00 & 237.00 & 7.31 & 12.50 & 0.28 & sc & 117.00 & 74.00 & 5.79 & 14.80 & 0.33 & trv & 177.00 & 95.00 & 6.01 & 8.93 & 0.39 \\
da & 146.00 & 92.00 & 6.14 & 9.44 & 0.25 & vep & 238.00 & 153.00 & 6.65 & 9.20 & 0.33 & mad & 74.00 & 52.00 & 5.42 & 10.33 & 0.33 \\
bg & 187.00 & 112.00 & 6.31 & 10.66 & 0.27 & kw & 54.00 & 39.00 & 5.10 & 8.00 & 0.25 & sm & 50.00 & 33.00 & 4.81 & 11.00 & 0.33 \\
et & 102.00 & 70.00 & 5.77 & 6.73 & 0.28 & kab & 51.00 & 37.00 & 4.95 & 7.86 & 0.25 & gcr & 152.00 & 86.00 & 5.99 & 14.44 & 0.31 \\
el & 280.00 & 162.00 & 6.79 & 13.43 & 0.27 & rw & 125.00 & 88.00 & 6.17 & 10.00 & 0.25 & pcm & 218.00 & 120.00 & 6.43 & 18.27 & 0.41 \\
hr & 157.00 & 100.00 & 6.15 & 9.30 & 0.26 & fiu-vro & 37.00 & 28.00 & 4.57 & 4.11 & 0.25 & gpe & 221.00 & 116.00 & 6.37 & 12.82 & 0.23 \\
lt & 128.00 & 83.00 & 5.89 & 9.00 & 0.30 & gv & 99.00 & 64.00 & 5.68 & 8.67 & 0.20 & pih & 31.00 & 23.00 & 4.32 & 6.11 & 0.33 \\
gl & 180.00 & 108.00 & 6.29 & 9.20 & 0.21 & mt & 462.00 & 210.00 & 6.67 & 18.10 & 0.30 & kcg & 75.00 & 54.00 & 5.58 & 13.00 & 0.25 \\
sl & 139.00 & 90.00 & 6.03 & 7.38 & 0.22 & frp & 54.00 & 43.00 & 5.26 & 4.67 & 0.24 & ss & 82.00 & 58.00 & 5.55 & 8.11 & 0.35 \\
nn & 122.00 & 80.00 & 6.00 & 9.00 & 0.25 & pcd & 93.00 & 65.00 & 5.67 & 5.67 & 0.24 & gur & 164.00 & 84.00 & 5.88 & 12.33 & 0.26 \\
ta & 116.00 & 85.00 & 6.10 & 7.60 & 0.23 & gn & 52.00 & 39.00 & 5.11 & 7.10 & 0.27 & ee & 36.00 & 29.00 & 4.69 & 7.00 & 0.33 \\
th & 262.00 & 142.00 & 6.69 & 13.53 & 0.03 & csb & 46.00 & 37.00 & 5.05 & 7.07 & 0.26 & chy & 9.00 & 8.00 & 2.88 & 2.00 & 0.25 \\
bn & 227.00 & 138.00 & 6.67 & 9.00 & 0.20 & smn & 89.00 & 61.00 & 5.52 & 4.73 & 0.20 & ik & 10.00 & 10.00 & 3.22 & 3.00 & 0.33 \\
lv & 135.00 & 88.00 & 6.01 & 9.15 & 0.26 & ay & 108.00 & 72.00 & 5.86 & 2.96 & 0.27 & fon & 92.00 & 58.00 & 5.52 & 8.58 & 0.32 \\
af & 162.00 & 95.00 & 5.97 & 8.44 & 0.20 & lez & 92.00 & 67.00 & 5.79 & 5.56 & 0.25 & ady & 41.00 & 33.00 & 4.87 & 6.00 & 0.33 \\
br & 89.00 & 59.00 & 5.57 & 7.33 & 0.29 & olo & 60.00 & 42.00 & 5.14 & 4.90 & 0.27 & guc & 100.00 & 68.00 & 5.80 & 14.59 & 0.38 \\
ml & 147.00 & 105.00 & 6.35 & 9.16 & 0.24 & mwl & 169.50 & 100.00 & 6.08 & 14.70 & 0.33 & fat & 241.00 & 125.00 & 6.39 & 18.95 & 0.33 \\
be-x-old & 101.00 & 71.00 & 5.84 & 8.37 & 0.30 & lfn & 176.00 & 92.00 & 5.95 & 17.40 & 0.35 & pwn & 198.00 & 89.00 & 5.75 & 16.45 & 0.36 \\
nds & 166.00 & 101.00 & 6.21 & 11.29 & 0.30 & kaa & 64.00 & 49.00 & 5.35 & 9.74 & 0.32 & din & 206.00 & 115.00 & 6.24 & 18.54 & 0.33 \\
lb & 106.00 & 67.00 & 5.72 & 7.50 & 0.25 & stq & 80.00 & 55.00 & 5.50 & 10.33 & 0.33 & ti & 27.00 & 23.00 & 4.38 & 7.00 & 0.33 \\
ga & 56.00 & 42.00 & 5.20 & 6.86 & 0.20 & ang & 51.00 & 39.00 & 5.11 & 9.00 & 0.33 & kl & 61.00 & 45.00 & 5.26 & 8.79 & 0.33 \\
is & 96.00 & 68.00 & 5.80 & 10.07 & 0.27 & fur & 68.00 & 51.00 & 5.47 & 7.83 & 0.30 & dz & 1008.50 & 214.50 & 4.82 & 86.37 & 0.54 \\
fy & 231.00 & 127.00 & 6.39 & 12.07 & 0.27 & ext & 90.00 & 63.00 & 5.70 & 13.00 & 0.33 & fi & 141.00 & 97.00 & 6.22 & 8.82 & 0.23 \\
pa & 186.00 & 111.00 & 6.39 & 13.54 & 0.24 & tw & 218.00 & 116.00 & 6.32 & 16.24 & 0.31 & sg & 11.00 & 10.00 & 3.26 & 2.33 & 0.11 \\
tl & 149.00 & 84.00 & 5.91 & 12.50 & 0.25 & lad & 117.00 & 74.00 & 5.84 & 9.14 & 0.20 & ts & 76.00 & 52.00 & 5.40 & 11.52 & 0.33 \\
an & 116.00 & 70.00 & 5.74 & 8.60 & 0.30 & gom & 319.00 & 207.00 & 7.00 & 13.00 & 0.31 & bm & 16.00 & 13.00 & 3.58 & 2.44 & 0.14 \\
wuu & 65.00 & 48.00 & 5.41 & 42.00 & 1.00 & pap & 102.00 & 63.00 & 5.61 & 11.00 & 0.29 & ny & 68.00 & 49.00 & 5.42 & 12.38 & 0.33 \\
sco & 108.00 & 66.00 & 5.68 & 7.18 & 0.18 & tyv & 238.00 & 143.00 & 6.57 & 9.87 & 0.33 & ltg & 60.00 & 44.00 & 5.21 & 6.86 & 0.33 \\
ha & 182.50 & 108.00 & 6.34 & 13.42 & 0.27 & ln & 33.00 & 25.00 & 4.46 & 5.08 & 0.25 & om & 75.50 & 56.00 & 5.55 & 15.55 & 0.33 \\
ne & 62.00 & 48.00 & 5.44 & 5.40 & 0.17 & ksh & 117.00 & 81.00 & 6.03 & 10.77 & 0.33 & inh & 47.00 & 37.00 & 5.03 & 5.90 & 0.20 \\
kn & 263.00 & 177.00 & 6.95 & 12.67 & 0.30 & bxr & 89.00 & 63.00 & 5.69 & 8.97 & 0.25 & dv & 78.00 & 58.00 & 5.63 & 10.97 & 0.37 \\
als & 296.00 & 165.00 & 6.76 & 11.42 & 0.30 & blk & 141.00 & 100.00 & 6.21 & 38.05 & 0.41 & dsb & 69.00 & 50.00 & 5.38 & 5.83 & 0.21 \\
bar & 106.00 & 72.00 & 5.86 & 6.79 & 0.25 & pdc & 37.00 & 28.50 & 4.68 & 6.00 & 0.33 & skr & 259.00 & 145.00 & 6.65 & 15.85 & 0.33 \\
ig & 310.00 & 154.00 & 6.57 & 16.87 & 0.29 & to & 61.00 & 42.00 & 5.11 & 10.40 & 0.33 & sat & 333.00 & 154.00 & 6.62 & 13.27 & 0.24 \\
si & 131.00 & 91.00 & 6.20 & 11.76 & 0.29 & tcy & 204.00 & 136.00 & 6.63 & 9.42 & 0.25 & km & 146.50 & 88.00 & 5.81 & 15.41 & 0.25 \\
frr & 53.00 & 39.00 & 5.08 & 6.67 & 0.20 & arc & 22.00 & 20.00 & 4.25 & 7.67 & 0.33 & mn & 158.00 & 104.00 & 6.34 & 8.62 & 0.21 \\
ps & 165.00 & 95.00 & 6.07 & 17.28 & 0.33 & mnw & 49.00 & 36.00 & 5.05 & 18.57 & 0.29 &  & & & & & \\
\bottomrule
\end{tabular}
    \caption{\footnotesize Heuristic values for Tier 1.}
\end{table*}

\begin{table*}[h]
    \centering
    \tiny
    \begin{tabular}{@{}lc cccc @{\hspace{5pt}} lc cccc @{\hspace{5pt}} lc cccc@{}}
\toprule
\textbf{Wiki-Id} & \textbf{Len.} & \textbf{Uniq.} & \textbf{Ent.} & \textbf{W/L} & \textbf{Pct.} & \textbf{Wiki-Id} & \textbf{Len.} & \textbf{Uniq.} & \textbf{Ent.} & \textbf{W/L} & \textbf{Pct.} & \textbf{Wiki-Id} & \textbf{Len.} & \textbf{Uniq.} & \textbf{Ent.} & \textbf{W/L} & \textbf{Pct.} \\
\midrule
nl & 52.00 & 38.00 & 5.06 & 10.00 & 0.33 & or & 166.00 & 104.00 & 6.32 & 8.32 & 0.20 & xh & 75.00 & 57.00 & 5.56 & 14.73 & 0.40 \\
uk & 161.00 & 105.00 & 6.30 & 7.67 & 0.28 & bcl & 113.00 & 69.00 & 5.73 & 14.00 & 0.29 & nia & 24.00 & 21.00 & 4.30 & 4.40 & 0.20 \\
fa & 71.00 & 48.00 & 5.34 & 4.39 & 0.18 & ilo & 76.00 & 46.00 & 5.11 & 8.07 & 0.19 & nov & 56.00 & 40.00 & 5.09 & 6.00 & 0.33 \\
id & 98.00 & 66.00 & 5.64 & 8.75 & 0.24 & am & 27.00 & 21.00 & 4.33 & 4.50 & 0.25 & ki & 19.00 & 16.00 & 3.93 & 5.33 & 0.33 \\
hu & 187.00 & 115.00 & 6.30 & 7.54 & 0.21 & sa & 72.00 & 50.00 & 5.38 & 3.14 & 0.14 & tn & 119.00 & 72.00 & 5.84 & 17.67 & 0.33 \\
ro & 74.00 & 48.00 & 5.29 & 6.15 & 0.20 & zu & 16.00 & 15.00 & 3.81 & 2.67 & 0.20 & kg & 16.00 & 11.00 & 3.38 & 5.36 & 0.33 \\
hy & 139.00 & 92.00 & 6.17 & 7.64 & 0.23 & hif & 39.00 & 29.00 & 4.67 & 5.17 & 0.18 & lbe & 18.00 & 17.00 & 4.00 & 3.67 & 0.33 \\
cy & 146.00 & 91.00 & 6.12 & 6.79 & 0.21 & mrj & 34.00 & 29.00 & 4.69 & 5.36 & 0.33 & ami & 131.50 & 61.00 & 5.34 & 9.30 & 0.35 \\
uz & 69.00 & 48.00 & 5.38 & 6.56 & 0.25 & ary & 123.00 & 71.00 & 5.81 & 7.48 & 0.24 & alt & 305.00 & 187.00 & 6.95 & 8.18 & 0.33 \\
sk & 81.00 & 57.00 & 5.57 & 6.40 & 0.20 & roa-tara & 21.00 & 18.00 & 4.12 & 4.00 & 0.25 & got & 24.00 & 20.00 & 4.19 & 4.56 & 0.25 \\
be & 101.00 & 71.00 & 5.84 & 8.37 & 0.30 & pam & 61.00 & 46.00 & 5.33 & 5.00 & 0.15 & rn & 21.00 & 18.00 & 3.94 & 4.29 & 0.33 \\
ur & 76.00 & 51.00 & 5.44 & 5.08 & 0.14 & bh & 74.00 & 48.00 & 5.23 & 7.63 & 0.20 & ch & 24.00 & 19.00 & 4.17 & 5.50 & 0.25 \\
az & 112.00 & 79.00 & 5.99 & 7.50 & 0.25 & myv & 43.00 & 33.00 & 4.93 & 3.83 & 0.25 & pnt & 68.00 & 51.00 & 5.44 & 4.75 & 0.21 \\
ka & 92.00 & 66.00 & 5.74 & 7.33 & 0.20 & se & 18.00 & 16.00 & 3.88 & 3.00 & 0.20 & iu & 14.00 & 12.00 & 3.46 & 4.26 & 0.33 \\
hi & 75.00 & 46.00 & 5.31 & 7.36 & 0.18 & bo & 122.00 & 47.00 & 3.98 & 22.33 & 0.60 & st & 20.00 & 18.00 & 4.17 & 15.00 & 0.44 \\
mk & 198.00 & 118.00 & 6.37 & 9.93 & 0.25 & tk & 19.00 & 17.00 & 4.04 & 3.67 & 0.20 & tpi & 17.00 & 14.00 & 3.77 & 4.33 & 0.33 \\
la & 71.00 & 54.00 & 5.54 & 5.15 & 0.15 & zea & 111.00 & 69.00 & 5.70 & 7.64 & 0.29 & anp & 71.00 & 50.00 & 5.35 & 7.00 & 0.18 \\
zh-yue & 48.00 & 34.00 & 4.91 & 7.20 & 0.22 & udm & 38.00 & 29.00 & 4.66 & 5.29 & 0.33 & krc & 31.00 & 27.00 & 4.70 & 3.20 & 0.20 \\
ast & 176.00 & 109.00 & 6.23 & 8.13 & 0.22 & kv & 65.00 & 44.00 & 5.08 & 4.70 & 0.25 & awa & 21.00 & 17.00 & 3.95 & 6.00 & 0.33 \\
my & 60.00 & 42.00 & 5.18 & 11.20 & 0.20 & nrm & 84.00 & 54.00 & 5.32 & 4.67 & 0.23 & av & 41.00 & 34.00 & 4.93 & 3.80 & 0.22 \\
sq & 80.00 & 57.00 & 5.49 & 7.32 & 0.20 & ks & 24.00 & 20.00 & 4.25 & 4.20 & 0.17 & nah & 18.00 & 15.00 & 3.79 & 1.82 & 0.14 \\
oc & 70.00 & 50.00 & 5.43 & 7.00 & 0.20 & lo & 87.00 & 61.00 & 5.60 & 12.02 & 0.16 & ug & 82.00 & 61.00 & 5.67 & 13.60 & 0.33 \\
sw & 62.00 & 42.00 & 5.14 & 5.80 & 0.17 & rm & 56.00 & 41.00 & 5.15 & 7.00 & 0.25 & rue & 56.00 & 43.00 & 5.21 & 7.25 & 0.25 \\
ky & 115.00 & 82.00 & 5.93 & 10.57 & 0.22 & dty & 39.00 & 30.00 & 4.74 & 3.40 & 0.15 & mhr & 60.00 & 39.00 & 4.93 & 6.67 & 0.30 \\
jv & 52.00 & 38.00 & 5.06 & 5.90 & 0.19 & lg & 61.00 & 45.00 & 5.18 & 7.33 & 0.33 & hsb & 116.00 & 83.00 & 5.97 & 6.70 & 0.28 \\
ba & 174.00 & 107.00 & 6.31 & 7.89 & 0.25 & za & 22.00 & 17.00 & 3.97 & 3.67 & 0.33 & os & 19.00 & 16.00 & 3.93 & 3.60 & 0.20 \\
io & 63.00 & 43.00 & 5.18 & 10.33 & 0.33 & gag & 27.00 & 24.00 & 4.50 & 2.10 & 0.20 & ckb & 66.00 & 46.00 & 5.32 & 5.29 & 0.17 \\
vo & 51.00 & 36.00 & 4.98 & 5.18 & 0.30 & szy & 181.00 & 88.00 & 5.78 & 8.04 & 0.33 & pnb & 89.00 & 51.00 & 5.33 & 4.91 & 0.18 \\
scn & 29.00 & 23.00 & 4.45 & 6.33 & 0.33 & tay & 125.00 & 69.00 & 5.70 & 5.60 & 0.43 & bs & 182.00 & 109.00 & 6.27 & 6.07 & 0.22 \\
qu & 65.00 & 46.00 & 5.26 & 3.40 & 0.25 & atj & 33.00 & 26.00 & 4.61 & 5.00 & 0.20 &  & & & & & \\
xmf & 72.00 & 53.00 & 5.48 & 6.80 & 0.18 & shi & 87.00 & 51.00 & 5.22 & 9.00 & 0.22 &  & & & & & \\
\bottomrule
\end{tabular}
    \caption{\footnotesize Heuristic values for Tier 2.}
\end{table*}

\begin{table*}[h]
    \centering
    \tiny

\begin{tabular}{@{}lc cccc @{\hspace{5pt}} lc cccc}
\toprule
\textbf{Wiki-Id} & \textbf{Len.} & \textbf{Uniq.} & \textbf{Ent.} & \textbf{W/L} & \textbf{Pct.} & \textbf{Wiki-Id} & \textbf{Len.} & \textbf{Uniq.} & \textbf{Ent.} & \textbf{W/L} & \textbf{Pct.}\\
\midrule
sv & 42.00 & 32.00 & 4.85 & 5.29 & 0.17 & bjn & 19.00 & 16.00 & 3.92 & 17.00 & 1.00 \\
arz & 51.00 & 36.00 & 4.98 & 2.46 & 0.18 & hak & 70.00 & 42.00 & 4.57 & 6.43 & 0.17 \\
vi & 40.00 & 32.00 & 4.93 & 4.33 & 0.14 & nso & 24.00 & 21.00 & 4.33 & 7.00 & 0.33 \\
sr & 70.00 & 51.00 & 5.48 & 3.76 & 0.15 & gan & 28.00 & 23.00 & 4.37 & 4.33 & 0.17 \\
sh & 70.00 & 51.00 & 5.48 & 3.76 & 0.15 & tly & 25.00 & 18.00 & 3.91 & 6.25 & 0.20 \\
eu & 51.00 & 41.00 & 5.19 & 4.50 & 0.20 & mdf & 53.00 & 39.00 & 5.03 & 2.09 & 0.15 \\
kk & 99.00 & 76.00 & 5.97 & 6.82 & 0.20 & koi & 52.00 & 40.00 & 4.92 & 4.21 & 0.25 \\
tg & 51.00 & 39.00 & 5.09 & 3.45 & 0.12 & cbk-zam & 64.00 & 46.00 & 5.27 & 2.80 & 0.12 \\
lmo & 55.00 & 38.00 & 5.01 & 6.36 & 0.20 & pfl & 90.00 & 57.00 & 5.53 & 3.92 & 0.30 \\
vec & 37.00 & 27.00 & 4.57 & 4.10 & 0.13 & haw & 24.00 & 18.00 & 4.02 & 7.00 & 0.33 \\
ht & 49.00 & 37.00 & 5.03 & 2.57 & 0.11 & ty & 17.00 & 12.00 & 3.50 & 5.00 & 0.33 \\
pms & 62.00 & 41.00 & 5.17 & 14.67 & 0.33 & srn & 56.00 & 38.00 & 5.02 & 9.05 & 0.20 \\
su & 31.00 & 23.00 & 4.47 & 11.73 & 0.29 & chr & 17.00 & 16.00 & 3.91 & 1.75 & 0.14 \\
szl & 36.00 & 29.00 & 4.71 & 8.50 & 0.25 & ve & 26.00 & 23.00 & 4.46 & 5.67 & 0.33 \\
diq & 47.00 & 36.00 & 4.95 & 2.71 & 0.14 & cr & 9.00 & 8.00 & 3.00 & 2.33 & 0.33 \\
yo & 10.00 & 9.00 & 3.12 & 1.60 & 0.20 & xal & 14.00 & 14.00 & 3.73 & 2.64 & 0.17 \\
ia & 30.00 & 22.00 & 4.32 & 9.00 & 0.33 & ab & 16.00 & 16.00 & 4.00 & 0.03 & 0.00 \\
gu & 93.00 & 63.00 & 5.69 & 19.00 & 0.33 & glk & 31.00 & 25.00 & 4.55 & 6.55 & 0.33 \\
bpy & 112.00 & 78.00 & 5.87 & 6.07 & 0.27 & dag & 75.00 & 52.00 & 5.27 & 2.25 & 0.07 \\
mzn & 53.00 & 40.00 & 5.19 & 5.89 & 0.17 & cdo & 21.00 & 16.00 & 3.73 & 6.00 & 0.33 \\
bat-smg & 21.00 & 19.00 & 4.20 & 5.67 & 0.33 & tum & 37.00 & 23.00 & 4.28 & 5.40 & 0.17 \\
nap & 39.00 & 18.00 & 3.76 & 5.14 & 0.14 & ban & 76.00 & 50.00 & 5.43 & 5.08 & 0.15 \\
gor & 29.00 & 18.00 & 4.03 & 3.71 & 0.14 & nv & 68.00 & 48.00 & 5.27 & 11.62 & 0.50 \\
mai & 49.00 & 36.00 & 5.11 & 3.30 & 0.11 & crh & 21.00 & 19.00 & 4.20 & 6.33 & 0.33 \\
map-bms & 52.00 & 38.00 & 5.06 & 5.90 & 0.19 & ku & 36.00 & 27.00 & 4.55 & 3.75 & 0.20 \\
shn & 103.00 & 53.00 & 5.48 & 8.40 & 0.20 & te & 341.00 & 211.00 & 7.02 & 10.42 & 0.25 \\
eml & 14.00 & 9.00 & 2.77 & 1.67 & 0.00 & mr & 89.00 & 58.00 & 5.45 & 5.50 & 0.25 \\
ace & 29.00 & 19.00 & 4.11 & 4.83 & 0.17 & mg & 58.00 & 35.00 & 4.97 & 3.53 & 0.20 \\
ie & 39.00 & 27.00 & 4.57 & 10.00 & 0.33 & ms & 47.00 & 32.00 & 4.79 & 4.70 & 0.18 \\
lij & 36.00 & 31.00 & 4.88 & 1.00 & 0.04 &  & & & & & \\
\bottomrule
\end{tabular}
    \caption{\footnotesize Heuristic values for Tier 3.}
\end{table*}

\begin{table*}[h]
    \centering
    \tiny

\begin{tabular}{@{}lc cccc}
\toprule
\textbf{Wiki-Id} & \textbf{Len.} & \textbf{Uniq.} & \textbf{Ent.} & \textbf{W/L} & \textbf{Pct.}\\
\midrule
war & 40.00 & 29.00 & 4.65 & 7.20 & 0.20 \\
ce & 88.00 & 64.00 & 5.65 & 4.76 & 0.18 \\
tt & 108.00 & 72.00 & 5.79 & 6.17 & 0.21 \\
azb & 49.00 & 38.00 & 5.10 & 4.67 & 0.20 \\
min & 57.00 & 47.00 & 5.46 & 5.64 & 0.25 \\
lld & 39.00 & 28.00 & 4.68 & 3.40 & 0.10 \\
new & 69.00 & 47.00 & 5.33 & 3.44 & 0.24 \\
cv & 105.00 & 72.00 & 5.82 & 5.32 & 0.18 \\
avk & 162.00 & 76.00 & 5.27 & 7.60 & 0.44 \\
mni & 27.00 & 23.00 & 4.46 & 4.00 & 0.14 \\
mi & 91.00 & 54.00 & 5.41 & 9.11 & 0.23 \\
pag & 58.00 & 42.00 & 5.17 & 3.38 & 0.06 \\
pi & 15.00 & 14.00 & 3.77 & 0.83 & 0.08 \\
bug & 19.00 & 12.00 & 3.38 & 3.33 & 0.17 \\
ceb & 88.00 & 46.00 & 5.11 & 10.08 & 0.27 \\
\bottomrule
\end{tabular}
    \caption{\footnotesize Heuristic values for Tier 4.}
\end{table*}

\end{document}